%% file: slam.tex
\documentclass{article}
\usepackage[preprint]{NEURIPS2026/neurips_2026}

\usepackage[utf8]{inputenc}
\usepackage[T1]{fontenc}
\usepackage{microtype}
\usepackage{hyperref}
\usepackage{hypernat}
\usepackage{url}
\usepackage{booktabs}
\usepackage{graphicx}
\usepackage{amsmath}
\usepackage{amssymb}
\usepackage{nicefrac}
\usepackage{xcolor}
\usepackage{subcaption}
\usepackage{multirow}
\usepackage{colortbl}
\usepackage{threeparttable}
\usepackage{tablefootnote}

\input{math_commands}

\definecolor{darkblue}{rgb}{0, 0, 0.5}
\hypersetup{colorlinks=true, citecolor=darkblue, linkcolor=darkblue, urlcolor=darkblue}

\newcommand{\slam}{\textsc{SLAM}}

\newcommand{\calz}{\hat{z}}

\title{\slam: Structural Linguistic Activation Marking for Language Models}

\author{%
  Fabrice Harel-Canada \quad Amit Sahai$^1$ \\[0.3em]
  \small \texttt{fabricehc@cs.ucla.edu} \quad \texttt{sahai@cs.ucla.edu} \\[0.5em]
  \normalsize $^1$University of California, Los Angeles
}

\begin{document}

\maketitle

% ─────────────────────────────────────────────────────────────────────────────
\begin{abstract}
LLM watermarks must be detectable without compromising text quality, yet most existing schemes bias the next-token distribution and pay for detection with measurable quality loss.                                                                           
We present \slam{} (\textbf{S}tructural \textbf{L}inguistic \textbf{A}ctivation \textbf{M}arking), a novel white-box watermarking scheme that sidesteps this cost by writing the mark into structural geometry rather than token frequencies: sparse autoencoders identify residual-stream directions encoding linguistic structure (e.g., voice, tense, clause order), and we causally steer those directions at generation time, leaving lexical sampling and semantics unconstrained. 
On Gemma-2 2B and 9B, \slam{} achieves $100\%$ detection accuracy with a quality cost of only $1$--$2$ reward points---compared to $7.5$--$11.5$ for KGW, EWD, and Unigram---with naturalness and diversity preserved at near-unwatermarked levels across both models. 
The trade-off is a complementary robustness profile: \slam{} resists word-level edits but is vulnerable to paraphrase that restructures syntax (at a quality cost), the converse of token-distribution methods.
\end{abstract}

\input{sections/01-intro}
\input{sections/02-background}

\input{sections/03-method}
\input{sections/04-experiments}
\input{sections/05-discussion}
\input{sections/06-conclusion}

% ─────────────────────────────────────────────────────────────────────────────
\section*{Acknowledgements}

We thank Professor Nanyun (Violet) Peng for generously providing compute resources that made this research possible.
This research was supported in part from a Simons Investigator Award, DARPA expMath award, Laude Moonshot grant, NSF grant 2333935, BSF grant 2022370, a Xerox Faculty Research Award, a Google Faculty Research Award, an Okawa Foundation Research Grant, and the Symantec Chair of Computer Science.
This material is based upon work supported by the Defense Advanced Research Projects Agency through Award HR001126CE054.

% ─────────────────────────────────────────────────────────────────────────────
\bibliographystyle{plainnat}
\bibliography{slam}

% ─────────────────────────────────────────────────────────────────────────────
\appendix

\input{sections/appendix/01-transferability}
\pagebreak
\input{sections/appendix/02-attack_validity}
\pagebreak
\input{sections/appendix/03-diversity}
\pagebreak
\input{sections/appendix/04-ablations}
\pagebreak
\input{sections/appendix/05-timing}
\pagebreak
\input{sections/appendix/06-implementation_details}
\pagebreak
\input{sections/appendix/07-attack_details}
\pagebreak

\newpage
\input{NEURIPS2026/checklist.tex}

\end{document}

%% file: math_commands.tex
\usepackage{amsmath,amsfonts,bm}

\def\eqref#1{equation~\ref{#1}}
\def\1{\bm{1}}

\DeclareMathAlphabet{\mathsfit}{\encodingdefault}{\sfdefault}{m}{sl}
\SetMathAlphabet{\mathsfit}{bold}{\encodingdefault}{\sfdefault}{bx}{n}

%% file: sections/01-intro.tex
% ─────────────────────────────────────────────────────────────────────────────
\section{Introduction}
\label{sec:intro}

The widespread deployment of large language models has created an urgent need
for reliable provenance attribution: distinguishing AI-generated text from
human-authored text, and tracing generated content to its source model.
Watermarking, which embeds an imperceptible but detectable signal into
generated text, is a promising approach, but existing methods face a
fundamental tension between detectability, quality, and robustness.

The dominant paradigm \citep{kirchenbauer2023watermark, kuditipudi2023robust,
zhao2023provable, synthid2024} biases the model's next-token distribution during
generation and tests for the resulting statistical regularity at detection time.
These schemes are efficient and theoretically grounded, but biasing the sampling
distribution comes at a measurable cost: KGW, EWD, and Unigram lose
$7.5$--$11.5$ reward points relative to the unwatermarked baseline on Gemma-2
PT models, and their conditional perplexity ratios collapse to $0.30$--$0.42$
(repetitive output the LM finds easier to predict than its own natural
generation, the signature of green-list-token recurrence; see
\S\ref{sec:experiments:main}). SAEMark \citep{saemark2025} uses SAEs as a post-hoc scorer over $N{=}50$
generation candidates, achieving near-lossless quality but at a significant
$O(N \times M)$ generation cost. Furthermore, its reliance on semantic features 
leaves it vulnerable to domain shift and erasure from simple text edits.

\paragraph{Our approach.}
\slam{} sidesteps token-distribution biasing entirely by writing the watermark
into the residual stream's structural geometry rather than into token
frequencies.
Syntactic structure is encoded in specific, localized subspaces of the LLM
residual stream \citep{park2023linear, hewitt2019structural}, a fact made
operational by public SAEs such as Gemma~Scope \citep{lieberum2024gemma}.
Unlike semantic features, structural features (passive voice, PP-fronting) are
by construction domain-invariant: a passive-voice SAE direction fires regardless
of whether the subject is a scientist or a banker.
\slam{} \emph{causally} steers those directions during generation, biasing
\emph{which} structural variant the model produces (e.g., passive vs.\ active
voice) without constraining \emph{which tokens} it samples.
The result is a watermark that leaves the lexical surface essentially
unconstrained. On reward quality, \slam{} achieves $\Delta\text{Reward}={-}1.3/{-}1.9$
on 2B/9B — best among all methods on 2B, and second only to SynthID on 9B
(SynthID's $-0.2$ comes paired with PPL ratio $0.72$, indicating repetitive
text). \slam{}'s PPL ratio of $1.24/1.36$ stays closest to $1$ on 2B, and
distinct-$n$, Self-BLEU, and MAUVE all remain within tight tolerance of the
unwatermarked baseline at a modest $\sim1.65\times$ overhead.

The trade-off is a complementary robustness profile: word-level edits leave
\slam{} intact ($100\%$ post-attack True Positive Rate (TPR) on synonym substitution, deletion,
and reordering) while syntax-restructuring paraphrase (e.g., DIPPER) can remove the signal
at a quality cost to the attacker, the converse of token-distribution methods.
Because we explicitly target provenance attribution rather than tamper-proof
cryptographic authentication, this vulnerability to severe structural
rewriting is a calculated, acceptable cost for near-lossless generation quality.

\paragraph{Contributions.}
\begin{enumerate}
  \item A contrastive mining pipeline (46,579 sentence pairs spanning 104
  syntactic, morphological, and discourse-level phenomena, each elicited
  across 5 semantic domains) that isolates structural SAE directions via
  composite scoring (contrastive $\times$ purity $\times$ cross-domain
  consistency).
  \item \textbf{PCA-bidirectional structural directions}: SVD of the contrastive
  difference matrix yields $k$ orthogonal structural modes per phenomenon,
  allowing \slam{} to operate on individual SAE latents ($k{=}1$) or composite
  subspaces ($k{>}1$); bidirectional mining captures both directions of each
  alternation, doubling the steerable bank. This handles distributed structural
  representations in larger models and enables a quality--detection trade-off
  (Figure~\ref{fig:k_alpha}).
  \item A complete white-box watermarking scheme with HMAC-keyed feature selection
  and an early-stop generation loop (cap $N\!\leq\!4$ candidates per prompt;
  the first candidate clearing the calibrated $\hat{z}$ threshold is returned),
  an order of magnitude fewer LLM calls than SAEMark.
  End-to-end overhead is $1.7\times$/$1.6\times$ the unwatermarked baseline on 2B/9B,
  with detection in one model forward pass ($\sim$420--640\,ms;
  vs.\ Adaptive's $\sim$10.7\,s, Table~\ref{tab:timing}).
  \item Empirical validation on Gemma~2 2B and 9B (PT) against six baselines
  across four families of quality metrics ($\Delta$Reward; conditional PPL
  ratio; grammar-error rate; distinct-$n$, Self-BLEU, MAUVE) at matched
  detection rate: \slam{} is the only method that does not catastrophically
  fail on any metric, paired with an attack-validity study establishing the
  complementary robustness profile (word-level immune, syntax-targeted-paraphrase-vulnerable).
\end{enumerate}

%% file: sections/02-background.tex
% ─────────────────────────────────────────────────────────────────────────────
\section{Background}
\label{sec:background}

\subsection{LLM watermarking}
\label{sec:background:watermarking}

% \footnote{Throughout, we target \emph{provenance attribution} (i.e., was this text generated
% by this model?) rather than tamper-proof cryptographic authentication; this
% scope shapes both the threat model and the trade-offs we accept.}

Token-distribution watermarks \citep{kirchenbauer2023watermark,
kuditipudi2023robust, zhao2023provable, synthid2024, hu2023unbiased} bias
next-token sampling (via green/red token lists, exponential minimum sampling, 
or entropy-aware selection) and detect the
resulting statistical regularity at inference time.
All produce signal in token-frequency space; we show this is disrupted by
moderate paraphrase (50--64\% TPR post-DIPPER for KGW/EWD/Unigram;
Figure~\ref{fig:robustness_heatmap}).

\paragraph{Non-distortion as a quality benchmark.}
\citet{synthid2024} formalise the gold standard for quality preservation as
\emph{$K$-sequence non-distortion}: over $K$ consecutive responses, the
marginal distribution of any single response equals the unwatermarked
distribution.
SynthID provably satisfies $K{=}1$ non-distortion; this aggregate-distributional
guarantee implies equal marginal distributions, so the measurable reward gap
we observe (\S\ref{sec:experiments:main}) likely reflects finite-sample variance
or the MarkLLM tournament-sampling implementation not perfectly realising the
theoretical ideal at $N{=}100$ prompts.
KGW, EWD, and Unigram make no such guarantee and pay a substantially
larger quality cost.
\slam{} does not prove non-distortion (steering modifies the intermediate
distribution rather than the sampling distribution) but empirically
approaches it on every quality axis we measure.

\paragraph{Linguistic and syntactic watermarking.}
An earlier line of work embeds watermarks through lexical and syntactic
rewriting: synonym substitution \citep{atallah2001natural, chang2010syntactic},
masked-LM edits \citep{ueoka2021frustratingly}, and invariant syntactic features
\citep{pmlr-v202-yoo23b}.
These surface-form signals are eliminated by any paraphraser not constrained to
preserve exact wording.
Semantic-invariant watermarks \citep{liu2023semantic, hou2023semstamp, ren2023semamark}
and post-hoc black-box marking \citep{chang2024postmark} seek robustness without
model access.
White-box functional-invariant marking \citep{fernandez2023functional} and
distribution-free schemes \citep{giboulot2024watermax} pursue robustness via
weight-space or generation-level invariance.

% \paragraph{Causal vs.\ output-syntactic.}
% All prior syntactic schemes operate on the generated text---picking among
% synonyms, applying syntactic transforms, or rewriting structure post-hoc.
% \slam{} operates on the residual stream that produces the text: the
% watermark is encoded in \emph{which structural variant} the model selects
% during sampling, not in any post-hoc rewrite. This is what makes lexical
% sampling unconstrained and is the source of the quality preservation we
% report in \S\ref{sec:experiments:main}.

% \paragraph{Theoretical limits and trade-offs.}
% \citet{christ2024undetectable} construct undetectable watermarks under
% cryptographic assumptions; \citet{zhang2023watermarks} establish impossibility
% results for certain strong-watermarking regimes.
% \citet{pang2024nofree} characterize fundamental quality--robustness--detectability
% trade-offs; our $k \times \alpha$ sweep and attack validity analysis
% (Section~\ref{sec:experiments:robustness}) provide empirical evidence for where
% \slam{} sits in this space.

\paragraph{SAEMark.}\label{sec:background:saemark}
SAEMark \citep{saemark2025}, the closest prior work, uses SAEs as a per-candidate
watermark scorer \emph{during} generation: at each of $M{=}10$ sentence steps it
samples $N{=}50$ candidate continuations, scores each via a Feature Concentration
Score (FCS) computed by an anchor SAE, and selects the candidate whose FCS ratio
best matches a secret pseudo-random target.
The result is generation that is ${\sim}12\times$ slower than \slam{} on equivalent
hardware (Table~\ref{tab:timing}).
The anchor model is recommended to be the Gemma-2 2B SAE regardless of generation model size.
\slam{} contrasts on four axes: (a) causal SAE use (steering, not scoring);
(b) structural features (domain-invariant by construction); (c) $O(N)$
generation cost with $N \leq 4$, requiring no candidate selection loop,
quality retry, or separate anchor model; and (d) robustness: SAEMark's
FCS depends on semantic feature activations that lexical edits disrupt,
collapsing TPR to 2--27\% on word-level attacks (Table~\ref{tab:robustness}),
whereas our structural approach is much more robust: \slam{} maintains
100\% TPR on all word-level attacks on both models we investigate.

\subsection{Sparse autoencoders and activation steering}
\label{sec:background:sae}

Residual stream representations in LLMs exhibit \emph{superposition}:
individual neurons represent linear combinations of many features
\citep{elhage2022toy}.
Sparse autoencoders (SAEs) disentangle this by learning a sparse, overcomplete
dictionary $W_{\text{dec}} \in \mathbb{R}^{n \times d}$ ($n \gg d$) such that
$h \approx W_{\text{dec}} \phi(h)$, where $\phi(h) \in \mathbb{R}^n_{\geq 0}$
is sparse \citep{sharkey2022taking, bricken2023monosemanticity, cunningham2023sparse}.
Individual SAE features correspond to interpretable concepts
\citep{bricken2023monosemanticity, templeton2024scaling, gao2024scaling}.
Activation steering, adding a direction to the residual stream, biases the
model toward downstream representations associated with that direction
\citep{turner2023activation, zou2023representation, meng2022locating}.
\emph{Contrastive activation addition} \citep{panickssery2023steering}, the
mean residual-activation difference between contrastive sets, underlies our
direction construction, motivated by the \emph{linear representation hypothesis}
\citep{park2023linear, park2024geometry}.
Syntactic phenomena specifically have been shown to occupy linearly-decodable
subspaces of the hidden state \citep{hewitt2019structural}. Gemma~Scope
\citep{lieberum2024gemma}, public SAEs for all layers and widths of Gemma~2,
surfaces these as discrete, sparse directions, enabling this pipeline without
custom SAE training.

%% file: sections/03-method.tex
% ─────────────────────────────────────────────────────────────────────────────
\section{Method}
\label{sec:method}

\begin{figure*}[t]
  \centering
  \includegraphics[width=\linewidth]{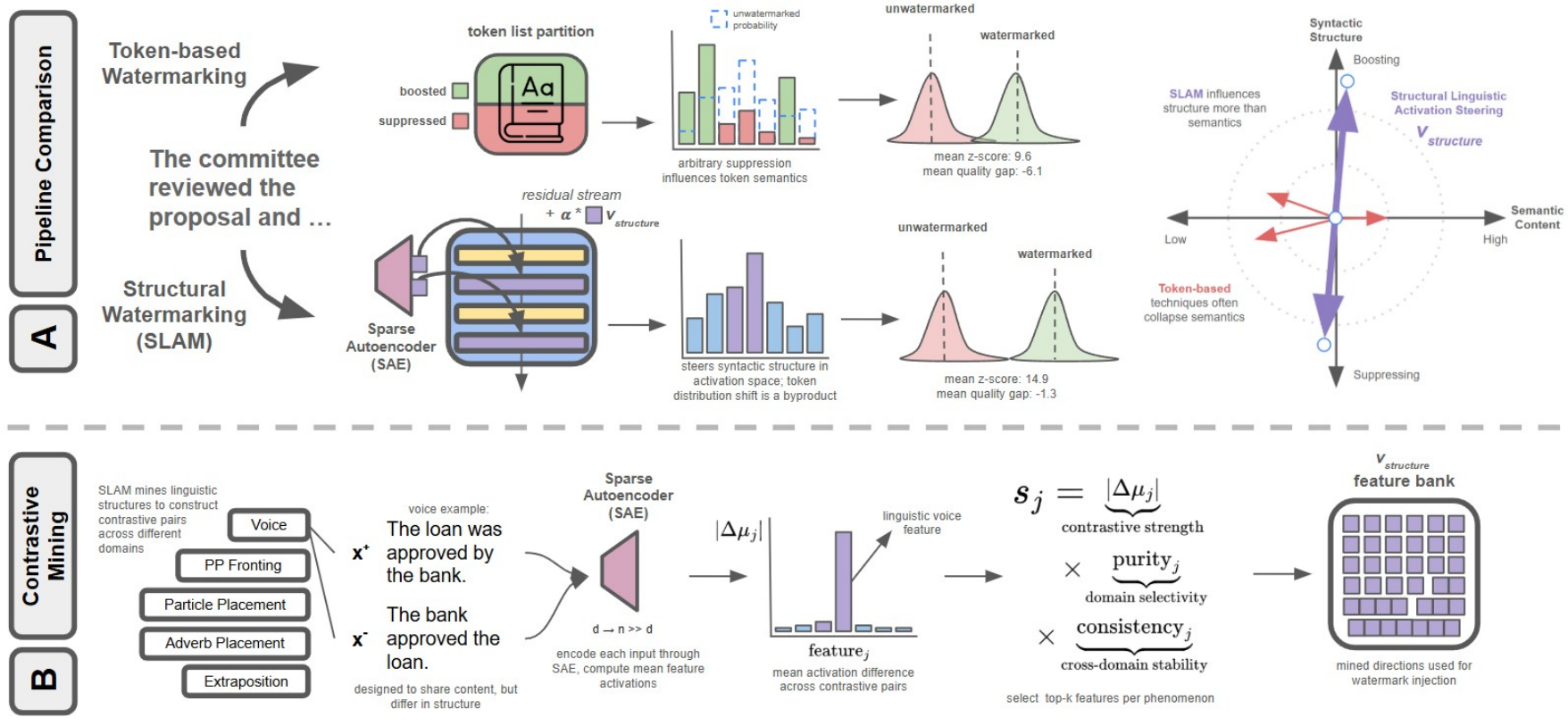}
  \caption{\textbf{Overview of \slam{}.}
    \textbf{(A)} Token-distribution watermarks bias token frequencies and pay
    for detection with measurable quality loss (\emph{top row}); \slam{}
    steers the residual stream along a structural direction, shifting syntactic
    form without distorting token semantics (\emph{bottom row}, geometric panel).
    \textbf{(B)} Contrastive sentence pairs isolate syntactic SAE features;
    SVD of the difference matrix yields $k$ orthogonal modes composed into the
    watermark direction $\mathbf{v}_{\text{structure}}$.}
  \label{fig:overview}
\vspace{-3mm}
\end{figure*}

\subsection{Contrastive dataset construction}
\label{sec:method:data}

We construct a dataset of 46,579 linguistically contrastive sentence pairs
covering 104 phenomena (filtered from a larger ${\sim}88$k-pair source pool to a
per-phenomenon cap during mining; full provenance in Appendix~\ref{app:impl:mining}).
Ninety phenomena are drawn from LinguaLens \citep{jing2025lingualens}; the
remaining 14 are hand-authored syntactic alternations derived from BLiMP
paradigms \citep{warstadt2020blimp}, targeting constructions particularly
relevant to watermarking such as passive voice, dative alternation, and cleft
constructions.
The combined set spans syntactic alternations, tense--aspect--mood marking,
morphological constructions, and discourse-level phenomena
(full list and per-phenomenon peak-layer heatmap: Appendix~\ref{app:impl:phenomena},
Figure~\ref{fig:phenomenon_layers}).
Each pair $(x^+, x^-)$ holds semantic content constant while varying only the
target construction.

Since BLiMP minimal pairs contrast grammatical with ungrammatical forms rather
than structurally polar equivalents, the 14 hand-authored phenomena were
constructed by generating semantically equivalent but structurally contrastive
pairs using Qwen3.5-9B \citep{qwen3} under phenomenon-specific instructions
with BLiMP vocabulary seeds for lexical diversity.
Generated pairs were then validated using AMRLib \citep{amrlib}, which
applies the AMR semantic framework \citep{banarescu2013abstract} to confirm
that $x^+$ and $x^-$ express the same underlying meaning despite different
surface forms.
Pairs span five semantic domains (finance, biology, sports, fiction, news),
preventing n-gram monotony and forcing features to generalize across content
domains; the cross-domain design ensures semantic variance cancels in the
mean, leaving only syntactic signal.
Mining proceeds per phenomenon rather than pooling all pairs, preserving
signal specificity: a feature responding only to passive voice would contribute
at ${\sim}1/104$ of its true contrastive strength when pooled across all 104 phenomena.

\subsection{Structural feature mining}
\label{sec:method:mining}

\paragraph{Per-token SAE encoding.}
For a text $x$ with tokens $t_1, \ldots, t_T$, we extract residual-stream
activations $h_l^{t_i} \in \mathbb{R}^d$ at layer $l$ and encode each token
individually through the SAE: $\phi_i = f(h_l^{t_i}) \in \mathbb{R}^n_{\geq 0}$.
We then mean-pool the sparse feature activations: $\bar{\phi}(x) =
\frac{1}{T}\sum_i \phi_i$.
Pooling residual-stream activations \emph{before} SAE
encoding is out-of-distribution (SAEs are trained on individual token
activations) and collapses positional structure that distinguishes structural
variants.

\paragraph{Contrastive scoring.}
For each SAE feature $j$, we compute the mean activation difference across the
contrastive dataset:
\begin{equation}
  \Delta\mu_j = \mathbb{E}_{(x^+, x^-) \sim \mathcal{D}}\bigl[\bar{\phi}_j(x^+) - \bar{\phi}_j(x^-)\bigr].
  \label{eq:contrastive}
\end{equation}
We additionally compute a \emph{purity} score (the fraction of pairs in
which feature $j$ activates more on $x^+$ than on $x^-$, measured on
C4 \citep{raffel2020c4} texts to assess in-the-wild selectivity) and a
\emph{consistency} score (the inverse coefficient of variation of
$\Delta\mu_j$ across the five semantic domains).
The composite score $s_j = |\Delta\mu_j| \times \text{purity}_j \times
\text{consistency}_j$ is used for candidate selection.
The composite filter passes roughly 2--4\% of raw candidates (full counts
in Appendix~\ref{app:impl:mining}), isolating structurally responsive
directions from features that merely co-vary with one or two domains.
Per-phenomenon banks are merged after trimming to a maximum bank size.

\subsection{Structural directions via SVD}
\label{sec:method:subspace}

\slam{} can watermark using individual SAE latents ($k{=}1$) or composite
subspaces ($k{>}1$).
In 2B-scale models, many structural concepts are represented by a handful of
co-activating SAE features, making $k{=}1$ effective.
In 9B-scale models with 16k-wide SAEs, the same concept is distributed across
$5$--$15$ weakly-discriminative features, so single-feature mining discards
most of the available signal and $k{>}1$ is necessary.

\slam{} constructs $k$ orthogonal composite directions via SVD of the
contrastive difference matrix.
Let $D \in \mathbb{R}^{N \times n}$ be the matrix of pair-wise activation
differences ($D_{ij} = \bar{\phi}_j(x_i^+) - \bar{\phi}_j(x_i^-)$).
We compute the thin SVD $D = U \Sigma V^\top$ and take the top-$k$ right
singular vectors $v_1, \ldots, v_k \in \mathbb{R}^n$ (sign-aligned so each
$v_j \cdot \Delta\mu > 0$):
\begin{equation}
  d_j = \frac{[v_j]_+ \cdot W_{\text{dec}}^\top}
           {\left\|[v_j]_+ \cdot W_{\text{dec}}^\top\right\|_2},
  \label{eq:pca}
\end{equation}
where $[v_j]_+$ clips negative components to zero and $W_{\text{dec}}$ maps
SAE-space directions back into the residual-stream.
Clipping to non-negative components is justified by the non-negativity of SAE
activations: negative entries of $v_j$ correspond to features whose
\emph{suppression} drives the contrast, an effect that cannot be enacted by
additive steering toward $x^+$.
Each $d_j$ is a unit vector in residual-stream space; all downstream operations
(steering, detection, z-scoring, calibration) are identical to the $k{=}1$ case.

\paragraph{Bidirectional mining.}
By default, \slam{} mines features where $x^+ > x^-$ (promotes the target
syntactic form).
With \emph{bidirectional} mining, we additionally process the reversed pairs
$(x^-, x^+)$, extracting directions that promote the \emph{base} form.
This doubles the per-phenomenon bank of steerable directions; the larger
pool matters because HMAC-keyed selection (\S\ref{sec:method:injection})
samples a per-document subset, and a larger pool reduces cross-document
collisions and increases per-document signal independence.
All results in this paper use bidirectional mining.

% \paragraph{Robustness conjecture.}
% An attacker must suppress an increasingly higher-dimensional subspace as
% $k$ grows; a systematic $k$-versus-direction-recovery study remains
% future work.

\subsection{Watermark injection}
\label{sec:method:injection}

As depicted in Figure~\ref{fig:overview}A, at inference time we hook into the
residual stream at layer $l$:
\begin{equation}
  h_l^{t} \leftarrow h_l^{t} + \alpha \cdot \mathbf{v}_{\text{structure}},
  \label{eq:steer}
\end{equation}
applied only to newly generated tokens (not the prompt prefix).
Each selected feature carries its own layer index, and injection is applied
per-feature at its respective layer; features targeting the same layer are
summed before the $\alpha$-scaled update is applied.
Phenomenon-specific peak layers span layers 2--14 on Gemma-2 2B and 2--22
on 9B, with morphological phenomena peaking earlier and syntactic alternations
later (Figure~\ref{fig:phenomenon_layers}; Appendix~\ref{app:impl:steer}).
The steering strength $\alpha$, the subspace dimension $k$, and the feature
bank size $f$ are initialized via parameter sweeps; final values and
sweep details are in Appendix~\ref{app:ablation}.
Larger $k$ achieves the same TPR at lower $\alpha$, yielding better quality
(Figure~\ref{fig:k_alpha}); final values are $\alpha{=}3$ (2B), $\alpha{=}12$ (9B).
A per-document feature subset is selected via
$\text{HMAC}(\text{key},\, \text{doc\_id})$-seeded strength-proportional
weighted sampling \citep{efraimidis2006weighted} over the feature bank.
We generate up to $N{=}4$ candidates and return the first clearing the
calibrated threshold ($\calz \geq 2.0$); if no candidate clears, the
highest-scoring candidate is returned as fallback.
Early-stop and degeneracy-filter details are in Appendix~\ref{app:impl:eval}.

\subsection{Detection and calibration}
\label{sec:method:detection}

Given a candidate text $x$, we run one model forward pass and project the
per-token residual-stream activations directly onto each pre-computed direction
$d_j \in \mathbb{R}^d$ (an $O(d)$ operation; no SAE forward pass is required at
detection time; see Appendix~\ref{app:timing}), then compute a raw Stouffer $z$-score:
\begin{equation}
  z_j = \frac{\bar{\phi}_j(x) - \mu_j^{\text{null}}}{\sigma_j^{\text{null}} / \sqrt{T}},
  \quad
  z_{\text{raw}} = \frac{\sum_{j \in \mathcal{A}} z_j}{\sqrt{|\mathcal{A}|}},
  \quad
  \mathcal{A} = \{j : z_j \geq 0.5\},
  \label{eq:zscore}
\end{equation}
where $\bar{\phi}_j(x) = \frac{1}{T}\sum_i \langle h_l^{t_i}, d_j \rangle$ denotes
the mean per-token projection of the residual stream onto direction $d_j$,
and $(\mu_j^{\text{null}}, \sigma_j^{\text{null}})$ are estimated on
unwatermarked model completions of the same source prompts used in
evaluation\footnote{Our technique does not require the unwatermarked source
prompts; a sufficiently general sample of texts from the model suffices to
approximate their structural signal strengths.}
(Appendix~\ref{app:impl:calib} details the calibration procedure).
Features whose direction-corrected $z_j$ falls below $0.5$ are dropped
before combining; this prevents noisy low-signal features from
diluting the combined statistic, and the divisor adapts to the number
of retained features rather than the bank size $k$.
We then calibrate the raw score against the null distribution:
\begin{equation}
  \calz = \frac{z_{\text{raw}} - \mu^{\text{null}}_{\text{raw}}}
               {\sigma^{\text{null}}_{\text{raw}}},
  \label{eq:calz}
\end{equation}
and threshold at $\calz \geq 2.0$, which corresponds to $\approx\!2.3\%$
False Positive Rate (FPR) under the null distribution's Gaussian fit.

%% file: sections/04-experiments.tex
% ─────────────────────────────────────────────────────────────────────────────
\section{Experiments}
\label{sec:experiments}

\subsection{Setup}
\label{sec:experiments:setup}

\paragraph{Models and SAEs.}
We evaluate on Gemma~2 2B and 9B pre-trained (PT) models \citep{gemmateam2024gemma2},
using Gemma~Scope \citep{lieberum2024gemma} residual-stream SAEs
(width 16k, $L_0 \approx 50$--80).
We use TransformerLens \citep{nanda2022transformerlens} for residual-stream
access and activation steering.
Transferability experiments (Appendix~\ref{app:transfer}) apply PT SAEs to the
corresponding instruction-tuned (IT) variants.

\paragraph{Baselines.}
We compare against six baselines: five token-distribution watermarks via
MarkLLM \citep{markllm}: KGW \citep{kirchenbauer2023watermark}
(green/red list bias), EWD \citep{kuditipudi2023robust} (entropy-weighted
detection over green-list bias; note that MarkLLM's EWD implementation uses
green-list bias rather than the distortion-free exponential minimum sampling
of the original paper), Unigram \citep{zhao2023provable}
(single-token green list), SynthID \citep{synthid2024} (tournament sampling
with repeated-context masking), and Adaptive \citep{liu2024adaptive}
(entropy-aware logit reweighting); and SAEMark \citep{saemark2025}, the
closest SAE-based prior work (\S\ref{sec:background:saemark}).

\paragraph{Evaluation prompts and metrics.}
We evaluate on a fixed 100-prompt set spanning five domains
(\emph{academic}, \emph{finance}, \emph{medicine}, \emph{news},
\emph{technology}; 20 prompts each). These domains are deliberately
disjoint from the contrastive-mining domains
(\S\ref{sec:method:data}: finance, biology, sports, fiction, news) on
three of five axes (finance and news appear in both), so feature mining cannot overfit to the eval
distribution.
Primary metrics: TPR at $\approx$2.3\% FPR (Gaussian upper-tail at
$\calz \geq 2$ under the per-bank null calibrated on $N\!=\!100$ baseline
texts; Appendix~\ref{app:impl:calib}), mean calibrated $z$-score
($\bar{\calz}$), and quality gap $\Delta\text{Reward} = \text{Reward}(\text{WM}) -
\text{Reward}(\text{BL})$ measured by Skywork-Reward-V2 \citep{skywork2024reward},
the top-ranked model on RewardBench \citep{lambert2025rewardbench}, a benchmark
for evaluating reward models used to assess AI-generated text quality.
Additional quality metrics (conditional PPL ratio, grammar-error rate,
distinct-$n$, Self-BLEU, MAUVE, and cosine-similarity drift) are introduced
as they appear in \S\ref{sec:experiments:main}.
Robustness is measured as post-attack TPR across seven attacks:
DIPPER \citep{krishna2024paraphrasing}, random walk (T5-XL span masking, 50 steps) \citep{zhang2023watermarks},
context synonym substitution, synonym substitution, word deletion, word
substitution, and sentence reordering as implemented in MarkLLM \citep{markllm}.
Per-attack implementation details (model, perturbation rate, target word
selection) are in Appendix~\ref{app:attacks}.

\subsection{Main results}
\label{sec:experiments:main}

Table~\ref{tab:main_results} reports TPR, $\bar{\calz}$, and $\Delta\text{Reward}$ for
all methods on both PT models. A $k \times \alpha$ ablation characterising \slam{}'s quality--detection
frontier is in Appendix~\ref{app:ablation}.

\input{table1_main_results}

\paragraph{Quality preservation.}
On Gemma~2 2B, \slam{} ($k{=}10$, PCA-bidirectional, $\alpha{=}3$) achieves 100\%
TPR at 2.3\% FPR with $\bar{\calz} = 14.95$, matching KGW, EWD, Unigram, and
Adaptive on detection while incurring a substantially smaller quality penalty
($\Delta\text{Reward} = {-}1.26$ vs.\ ${-}9.11$, ${-}8.57$, ${-}10.71$ for
KGW, EWD, Unigram).
\slam{}'s quality is competitive with SynthID ($-2.82$) and Adaptive ($-1.77$),
both of which use different mechanisms.
SAEMark achieves 99\% TPR on 2B ($\bar{\calz}=5.81$) with the strongest
single-axis quality score of any method ($\Delta\text{Reward}={-}0.59$,
PPL ratio $1.02$), but collapses to 13\% TPR on 9B, confirming that its
2B-tuned SAE features do not transfer to the larger model.
On Gemma~2 9B, \slam{} achieves the same pattern relative to the
token-distribution baselines ($\Delta\text{Reward} = {-}1.86$ vs.\
${-}7.49$ to ${-}11.53$ for KGW/EWD/Unigram); SynthID's
$\Delta\text{Reward} = {-}0.2$ is the only 9B result that
beats \slam{} on reward alone, but it comes paired with PPL ratio $0.72$,
which the naturalness analysis below places in the same predictability-collapse
band as the token-distribution baselines despite SynthID's distinct
tournament-sampling mechanism.

\paragraph{Output naturalness and distribution.}
Conditional PPL ratio (defined in Table~\ref{tab:main_results} and
Appendix~\ref{app:impl:ppl}) gives an independent quality signal:
\slam{} attains $1.24/1.36$ on 2B/9B (closest to 1 on 2B; on 9B
SynthID's $0.72$ is closer). The KGW/EWD/Unigram band of
$0.30$--$0.42$ indicates collapse onto repetitive, easy-to-predict
text; Adaptive over-shoots to $1.89$--$2.19$.
\slam{} also preserves the natural output distribution across four
further measures
(Tables~\ref{tab:semantic_sim}~and~\ref{tab:diversity},
Appendix~\ref{app:diversity}):
distinct-$n$ uniquely closest to baseline ($|\Delta|\!=\!0.013/0.024$;
KGW/EWD/Unigram lose $\geq\!50\%$ of unique bigrams), Self-BLEU within
$|\Delta|\!=\!0.015/0.017$ of baseline, MAUVE top on 9B (0.53) and
second on 2B (0.19; SAEMark leads at 0.90 but collapses to 0.007 on 9B),
and smallest semantic drift under Qwen3-Embedding-8B
cosine distance ($\Delta\!=\!+0.035/+0.047$; every token-distribution
baseline $\geq\!+0.045$).
Steering \emph{which} syntactic form the model produces, not \emph{which tokens}
it samples, leaves both lexical surface and meaning essentially unconstrained.

\paragraph{Grammar.}
\slam{}'s grammar error rate sits at the unwatermarked-baseline
level on 2B (1.33 vs.\ 1.39 per 100 tokens) and slightly elevated on 9B
(1.55 vs.\ 1.31), measured using LanguageTool \citep{languagetool}.
KGW and Adaptive are the worst on this axis (2.02--2.90), Unigram the
lowest (0.98--1.02) but at the cost of an extreme distinct-$n$ collapse,
a vocabulary so constrained that LanguageTool's productive-form detectors
find little to fire on.
SAEMark's grammar rate (1.48/1.82 on 2B/9B) is above baseline (1.39/1.31),
consistent with the vocabulary pressure from restricting token selection
to high-SAE-activation candidates.

\paragraph{Quality evaluation as multi-axis triangulation.}
Single-metric quality oracles are unreliable \citep{harel-canada-etal-2025-sandcastles};
we triangulate across $\Delta$Reward, conditional PPL ratio, grammar-error
rate, and four distribution-distance measures.
\slam{} is the only method that does not catastrophically fail on any axis:
KGW fails on PPL ratio, grammar, and MAUVE; EWD and Unigram on PPL ratio
and distinct-$n$; Adaptive on PPL ratio and grammar; SynthID on
$\Delta$Reward and PPL ratio (2B).

\subsection{Robustness}
\label{sec:experiments:robustness}

\begin{figure}[t]
  \centering
  \includegraphics[width=\linewidth]{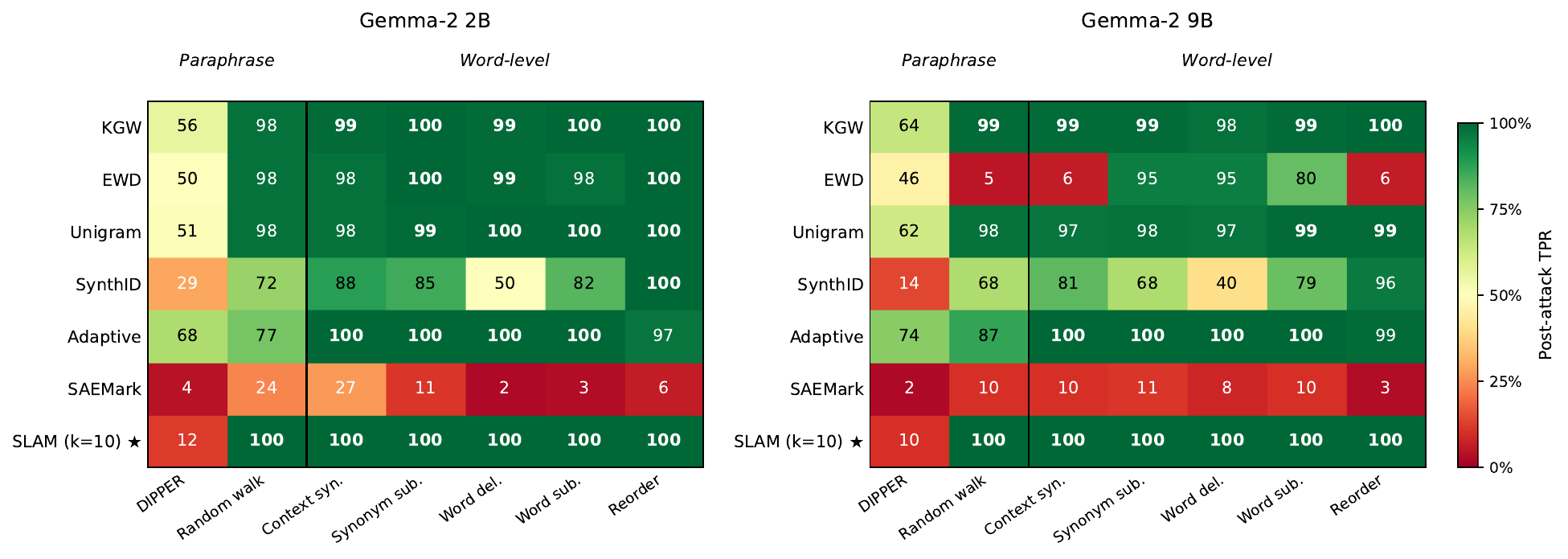}
  \caption{Post-attack TPR (\%) per method $\times$ attack on Gemma~2 2B
  (left) and 9B (right). Attacks are grouped by class
  (\emph{Paraphrase} / \emph{Word-level}); cells coloured by TPR
  (greener = more robust). \slam{} ($k{=}10$ PCA-bidirectional, starred
  row) is robust to all word-level attacks but vulnerable to
  syntax-restructuring paraphrase.}
  \label{fig:robustness_heatmap}
  \vspace{-3mm}
\end{figure}

\slam{}'s quality advantage comes from writing the watermark into
syntactic geometry rather than token frequencies, and the cost is
where you would expect: attacks that explicitly perturb syntax remove
the signal more easily than they would a token-distribution one.
Figure~\ref{fig:robustness_heatmap} reports post-attack TPR for each
method $\times$ attack on Gemma~2 2B and 9B. Per-attack averages are in Table~\ref{tab:robustness}.

\input{table3_robustness}

\slam{} is completely robust to all word-level and span-masking attacks
(100\% TPR on both models) but vulnerable to neural paraphrase that
restructures syntax: DIPPER reduces TPR to 12\%/10\% on 2B/9B, the
expected failure mode when the signal lives in syntactic geometry.

Token-distribution methods show the opposite profile: robust to word-level
attacks (KGW: $\geq\!98\%$ TPR) but eroded by neural paraphrase (KGW drops
to $56\%$/$64\%$ on DIPPER; Unigram to $51\%$/$62\%$).
Random walk (the technique-agnostic paraphraser) leaves \slam{} at
100\%/100\% TPR, matching the strongest baselines.
Adaptive is the most paraphrase-robust baseline (Avg(Para) $73\%/80\%$;
DIPPER $68\%/74\%$) at the cost of the quality degradation discussed in
\S\ref{sec:experiments:main}, illustrating the quality--paraphrase-robustness frontier.
SAEMark (2B) shows low post-attack TPR across all attacks (2--27\%),
reflecting that each attack further degrades the already-weak SAE-feature
signal even from a 99\% baseline.

%% file: table1_main_results.tex
\begin{table}[t]
\centering
\caption{Watermark detection performance and quality. TPR at $\approx$2.3\% FPR; Cal-Z = mean calibrated z-score; $\Delta$Reward = reward\textsubscript{WM} $-$ reward\textsubscript{BL} (Skywork-Reward-V2); PPL ratio = conditional perplexity on continuation tokens, watermarked / unwatermarked, scored by the generating model itself (see Appendix~\ref{app:impl}); Gram.\ Err.\ = LanguageTool grammar errors per 100 words. Arrows indicate direction of preference; \textbf{bold} = best per column within each panel. Top panel: PT models (main). Bottom panel: IT models (transferability).}
\label{tab:main_results}
\resizebox{\columnwidth}{!}{%
\begin{tabular}{llrrrrrrr}
\toprule
Model & Method & $N$ & TPR\,(\%)$\uparrow$ & Cal-Z$\uparrow$ & $\Delta$Reward$\uparrow$ & PPL ratio$\to1$ & Gram.\ Err.$\downarrow$ \\
\midrule
\multicolumn{8}{l}{\textit{Pre-trained models (PT SAE $\rightarrow$ PT model)}} \\
\midrule
  \multirow{7}{*}{Gemma-2 2B} & KGW & 100 & \textbf{100} & 10.2 & -9.1 & 0.37 & 2.47 \\
   & EWD & 100 & \textbf{100} & 6.8 & -8.6 & 0.35 & 1.34 \\
   & Unigram & 100 & \textbf{100} & 14.3 & -10.7 & 0.30 & \textbf{0.98} \\
   & SynthID & 100 & 99 & 7.2 & -2.8 & 0.62 & 1.26 \\
   & Adaptive & 100 & \textbf{100} & 10.6 & -1.8 & 1.89 & 2.27 \\
   & SAEMark & 100 & 99 & 5.8 & \textbf{-0.6} & \textbf{1.02} & 1.48 \\
   & SLAM (ours) & 100 & \textbf{100} & \textbf{14.9} & -1.3 & 1.24 & 1.33 \\
\midrule
  \multirow{7}{*}{Gemma-2 9B} & KGW & 100 & \textbf{100} & 11.4 & -7.5 & 0.41 & 2.02 \\
   & EWD & 100 & 99 & 5.5 & -7.7 & 0.41 & \textbf{0.86} \\
   & Unigram & 100 & \textbf{100} & \textbf{14.5} & -11.5 & 0.35 & 1.02 \\
   & SynthID & 100 & 99 & 7.2 & -0.2 & \textbf{0.72} & 1.24 \\
   & Adaptive & 100 & \textbf{100} & 10.7 & -2.4 & 2.19 & 2.90 \\
   & SAEMark & 100 & 13 & -1.3 & \textbf{+0.4} & 0.59 & 1.82 \\
   & SLAM (ours) & 100 & \textbf{100} & 11.8 & -1.9 & 1.36 & 1.55 \\
\midrule
\multicolumn{8}{l}{\textit{Instruction-tuned models (PT SAE $\rightarrow$ IT model)}} \\
\midrule
  Gemma-2 2B-IT & SLAM (ours) & 100 & 99 & 5.6 & -4.1 & 1.67 & 0.80 \\
  Gemma-2 9B-IT & SLAM (ours) & 100 & 93 & 11.3 & -4.1 & 1.69 & 0.63 \\
\bottomrule
\vspace{-7mm}
\end{tabular}}
\end{table}

%% file: table3_robustness.tex
\begin{table}[t]
\vspace{-2mm}
\centering
\caption{Post-attack TPR (\%) per method $\times$ attack on PT models.
``None'' = no-attack baseline.
\textbf{Avg(All)} = mean over all seven attacks;
\textbf{Avg(Para)} = mean over the two paraphrase attacks (DIPPER, Random Walk).
{\small $\bigstar$} = proposed SLAM ($k=10$);
\textbf{bold} = column max (ties bolded).}
\label{tab:robustness}
\resizebox{0.8\textwidth}{!}{%
\begin{tabular}{lrrrrrrrrrr}
\toprule
 &  & \multicolumn{2}{c}{\textit{Paraphrase}} & \multicolumn{5}{c}{\textit{Word-level}} & \multicolumn{2}{c}{\textit{Average}} \\
\cmidrule(lr){3-4} \cmidrule(lr){5-9} \cmidrule(lr){10-11}
Method & None & \rotatebox{90}{DIPPER} & \rotatebox{90}{Random walk} & \rotatebox{90}{Context syn.} & \rotatebox{90}{Synonym sub.} & \rotatebox{90}{Word del.} & \rotatebox{90}{Word sub.} & \rotatebox{90}{Reorder} & All & Para \\
\midrule
\multicolumn{11}{l}{\textit{Gemma-2 2B}}\\
\midrule
  KGW & \textbf{100} & 56 & 98 & 99 & \textbf{100} & 99 & \textbf{100} & \textbf{100} & \textbf{93} & \textbf{77} \\
  EWD & \textbf{100} & 50 & 98 & 98 & \textbf{100} & 99 & 98 & \textbf{100} & 92 & 74 \\
  Unigram & \textbf{100} & 51 & 98 & 98 & 99 & \textbf{100} & \textbf{100} & \textbf{100} & 92 & 74 \\
  SynthID & 99 & 29 & 72 & 88 & 85 & 50 & 82 & \textbf{100} & 72 & 50 \\
  Adaptive & \textbf{100} & \textbf{68} & 77 & \textbf{100} & \textbf{100} & \textbf{100} & \textbf{100} & 97 & 92 & 73 \\
  SAEMark & 99 & 4 & 24 & 27 & 11 & 2 & 3 & 6 & 11 & 14 \\
  SLAM (k=10) $\bigstar$ & \textbf{100} & 12 & \textbf{100} & \textbf{100} & \textbf{100} & \textbf{100} & \textbf{100} & \textbf{100} & 87 & 56 \\
\midrule
\multicolumn{11}{l}{\textit{Gemma-2 9B}}\\
\midrule
  KGW & \textbf{100} & 64 & 99 & 99 & 99 & 98 & 99 & \textbf{100} & \textbf{94} & \textbf{82} \\
  EWD & 99 & 46 & 5 & 6 & 95 & 95 & 80 & 6 & 48 & 26 \\
  Unigram & \textbf{100} & 62 & 98 & 97 & 98 & 97 & 99 & 99 & 93 & 80 \\
  SynthID & 99 & 14 & 68 & 81 & 68 & 40 & 79 & 96 & 64 & 41 \\
  Adaptive & \textbf{100} & \textbf{74} & 87 & \textbf{100} & \textbf{100} & \textbf{100} & \textbf{100} & 99 & \textbf{94} & 80 \\
  SAEMark & 13 & 2 & 10 & 10 & 11 & 8 & 10 & 3 & 8 & 6 \\
  SLAM (k=10) $\bigstar$ & \textbf{100} & 10 & \textbf{100} & \textbf{100} & \textbf{100} & \textbf{100} & \textbf{100} & \textbf{100} & 87 & 55 \\
\bottomrule
\vspace{-9mm}
\end{tabular}}
\end{table}

%% file: sections/05-discussion.tex
% ─────────────────────────────────────────────────────────────────────────────
\section{Discussion}
\label{sec:discussion}

\slam{} demonstrates that quality and detection need not trade off when
the watermark is encoded at the right level of linguistic abstraction.
The quality advantage traces to a single mechanistic cause: steering
\emph{which} structural variant the model produces leaves lexical sampling
entirely unconstrained, which is precisely what all four quality axes
independently confirm.

\paragraph{Why structural watermarks are quality-neutral by construction.}
Quality metrics measure properties of the output token sequence: reward models
score semantic coherence and fluency; conditional PPL measures how surprising
the generated tokens are to the model; grammar checkers flag surface-form errors.
None of these metrics are sensitive to \emph{which syntactic form} the model used
to express a given meaning---passive vs.\ active voice produces equally fluent,
equally coherent output with approximately equal surprisal to the generating model.
(In practice, PPL ratios of $1.24$/$1.36$ are far closer to the ideal of $1.0$
than the $0.30$--$0.42$ of token-distribution methods.)
\slam{}'s watermark carrier therefore lives in a subspace orthogonal, by
construction, to the axes that quality metrics inspect---under matched deployment
conditions (PT SAE on PT model); applying PT SAEs to IT models can shift prose
register in ways reward models penalise (Appendix~\ref{app:transfer}),
the dominant source of the quality penalty in that setting.

\paragraph{Multi-axis evaluation exposes single-metric blind spots.}
The SynthID 9B result illustrates why multi-axis measurement matters.
SynthID achieves $\Delta\text{Reward} = {-}0.2$ on 9B, suggesting near-perfect
quality preservation, yet its conditional PPL ratio is $0.72$---placing it in the
same repetition-collapse band as KGW and Unigram.
This divergence reveals a structural limitation of reward models as quality oracles:
they score semantic coherence and helpfulness well, but are insensitive to the
distributional regularities (repetitive token choices, restricted vocabulary range)
that conditional PPL and distinct-$n$ detect.
A watermark that satisfies a reward model while collapsing output entropy has not
achieved non-distortion; it has found a blind spot in the oracle.
Using four independent quality axes reduces this ambiguity: the methods that appear
problematic on any one axis are those that fail on at least two, and the methods
that appear uniformly good---\slam{}, and SynthID on 2B---are those that the
multi-axis picture consistently supports.

% \paragraph{Robustness frontier.}
% The complementary attack profiles, \slam{} immune to word-level attacks
% and vulnerable to syntax-targeted paraphrase while token-distribution
% methods show the converse, suggest ensemble deployment to cover both
% attack classes. \slam{}'s contribution is to push the quality axis
% without sacrificing word-level robustness; Adaptive provides stronger
% paraphrase coverage at the cost of quality and larger computation overhead.

% \paragraph{Mechanism: subspace dimension matters at scale.}
% At 9B, $k{=}1$ degrades to 90\% TPR even at $\alpha{=}20$, consistent
% with increasing superposition \citep{elhage2022toy, templeton2024scaling};
% $k{=}10$ recovers 100\%. Composing $k$ orthogonal modes into one
% steerable direction is the contribution that makes the scheme work at
% larger scale where structural concepts are distributed across
% many weakly-discriminative SAE features.

\paragraph{Limitations.}

\slam{} requires white-box residual-stream access;
detection is GPU-only, unlike KGW/Unigram/SynthID's CPU-side token scan.
DIPPER reduces TPR to 12\%/10\%; unlike token-distribution baselines where DIPPER
\emph{improves} quality by removing token-bias constraints
(e.g.\ KGW $\Delta\text{Reward}{=}{+4.4}/{+2.7}$), it imposes a small quality
cost on \slam{} ($\Delta\text{Reward}{=}{-1.1}/{-1.4}$ on 2B/9B;
Appendix~\ref{app:attack_validity}).
Paraphrase robustness can be partially
recovered by increasing $F$ at a quality cost (Appendix~\ref{app:inject_f}).
The scheme is single-bit; multi-bit extension via independent phenomena
is straightforward but not implemented.
Applying PT SAEs to IT models incurs a quality penalty
($\sim{-}4$ vs.\ $\sim{-}1.5$ $\Delta$Reward), motivating matched-SAE
deployment.
Lastly, our quality evidence is fully automated (four
metrics across $n\!=\!100$ prompts per condition); we have not run a
SynthID-scale human side-by-side preference study,
which would substantially strengthen the non-distortion claim.

\paragraph{Future work.}
Immediate priorities include deploying matched instruction-tuned SAEs to
reduce the quality penalty on IT models (Appendix~\ref{app:transfer}) and
extending \slam{} to a second model family using the recently released
Qwen-Scope SAEs \citep{qwen_scope}, which would also enable a cross-lingual
study of whether English-mined directions generalize within a multilingual model.
Mining augmentation with additional syntax-targeted phenomena is a
natural path toward narrowing the DIPPER vulnerability, since phenomena
with stronger paraphrase resistance (e.g., relative clause constructions)
would require more aggressive restructuring to evade detection.
Longer-term directions include multi-bit watermarking via independent phenomena,
multi-layer detection via crosscoder representations, and a deployment-scale
human preference study.

%% file: sections/06-conclusion.tex
% ─────────────────────────────────────────────────────────────────────────────
\section{Conclusion}
\label{sec:conclusion}

Encoding a watermark at the level of syntactic structure rather than token
frequencies breaks the coupling between detectability and quality.
\slam{} achieves this by causally steering structural SAE directions in the
residual stream, biasing \emph{which} syntactic form the model produces without
constraining \emph{which tokens} it samples.
The result is 100\% TPR at 2.3\% FPR on both Gemma-2 2B and 9B with a quality
gap ($\Delta\text{Reward}\,{=}\,{-}1.3/{-}1.9$) that is three to six times
smaller than the KGW, EWD, and Unigram baselines, paired with PPL ratio, distinct-$n$,
Self-BLEU, and MAUVE scores that no existing method simultaneously matches.
Composing $k{=}10$ orthogonal SVD modes is what makes this scale to 9B, where
distributed structural representations would otherwise defeat a single-direction
approach.
The trade-off is a complementary robustness profile: complete immunity to
word-level attacks and vulnerability to syntax-targeted paraphrase, the converse
of token-distribution methods, clarifying where each paradigm is appropriate
and motivating mining augmentation with more paraphrase-resistant phenomena as
the path toward closing the gap.

%% file: sections/appendix/01-transferability.tex
\raggedbottom
\section{Transferability to instruction-tuned models}
\label{app:transfer}

Table~\ref{tab:transferability} evaluates \slam{} with PT SAEs applied to the
corresponding IT variants.

\paragraph{2B-IT.}
Detection transfers cleanly: 99\% TPR at $\bar{\calz} = 5.62$
(vs.\ 14.95 on PT). Quality cost rises by $\sim\!2.8$ reward points
($\Delta\text{Reward} = {-}4.09$ vs.\ ${-}1.26$ on PT): PT structural
directions steer the IT model toward PT-style prose patterns,
which the reward model penalises. We hypothesise this is a
mismatch artifact that matched IT SAEs would recover; the experiment
to confirm is left to future work.
Robustness mirrors PT under syntax-targeted paraphrase
(DIPPER $6\%$ on IT vs.\ $12\%$ on PT 2B---lower robustness on IT
under the attack \slam{} is most vulnerable to).
Random walk and word-level attacks remain at $\geq\!93\%$.

\paragraph{9B-IT.}
\slam{} reaches 93\% TPR at $\bar{\calz} = 11.28$---lower TPR than the
PT baseline (100\%), which we hypothesise reflects the larger
representation shift induced by instruction tuning at this scale (matched
IT SAEs would discriminate this from a calibration issue). Quality
cost is similar to 2B-IT ($\Delta\text{Reward} = {-}4.08$).
Robustness numbers track the same pattern as 2B-IT but at higher base:
DIPPER $32\%$, synonym substitution $91\%$, word deletion $97\%$, random walk $93\%$.
The 93\% TPR with mismatched PT SAEs is a positive finding---the
watermark direction partially survives fine-tuning---without yet
being a deployment-ready setting.

\input{table4_transferability}

%% file: table4_transferability.tex
\begin{table}[!htbp]
\centering
\caption{PT SAE transferability to instruction-tuned models.
PT rows are reference values (from Table~\ref{tab:robustness}). IT rows apply the same PT SAE to the instruction-tuned variant.
Detection columns: TPR (\%), Cal-Z, $\Delta$Reward.
Robustness columns: post-attack TPR (\%) on all attacks.
\textbf{Bold} = 99--100\%.}
\label{tab:transferability}
\resizebox{\textwidth}{!}{%
\begin{tabular}{lrrr | rrrrrrrr}
\toprule
 & & & & & \multicolumn{2}{c}{\textit{Paraphrase}} & \multicolumn{5}{c}{\textit{Word-level}} \\
\cmidrule(lr){6-7}\cmidrule(lr){8-12}
Model & TPR & Cal-Z & $\Delta$Rew & None & \rotatebox{90}{DIPPER} & \rotatebox{90}{Random walk} & \rotatebox{90}{Context syn.} & \rotatebox{90}{Synonym sub.} & \rotatebox{90}{Word del.} & \rotatebox{90}{Word sub.} & \rotatebox{90}{Reorder} \\
\midrule
  Gemma-2 2B (PT ref.) & \textbf{100} & 14.9 & -1.3 & \textbf{100} & 12 & \textbf{100} & \textbf{100} & \textbf{100} & \textbf{100} & \textbf{100} & \textbf{100} \\
  Gemma-2 2B-IT & \textbf{99} & 5.6 & -4.1 & 99 & 6 & 98 & 95 & 94 & 99 & 100 & 99 \\
\midrule
  Gemma-2 9B (PT ref.) & \textbf{100} & 11.8 & -1.9 & \textbf{100} & 10 & \textbf{100} & \textbf{100} & \textbf{100} & \textbf{100} & \textbf{100} & \textbf{100} \\
  Gemma-2 9B-IT & 93 & 11.3 & -4.1 & 93 & 32 & 93 & 96 & 91 & 97 & 96 & 92 \\
\bottomrule
\end{tabular}}
\end{table}

%% file: sections/appendix/02-attack_validity.tex
\section{Attack quality impact}
\label{app:attack_validity}

This appendix expands on the attack-validity framing of
\S\ref{sec:experiments:robustness} along the quality axis: how much
each attack degrades the underlying text, so attacks that evade detection
by destroying the text are undermined as practical threats.

\begin{figure}[!htbp]
  \centering
  \includegraphics[width=\linewidth]{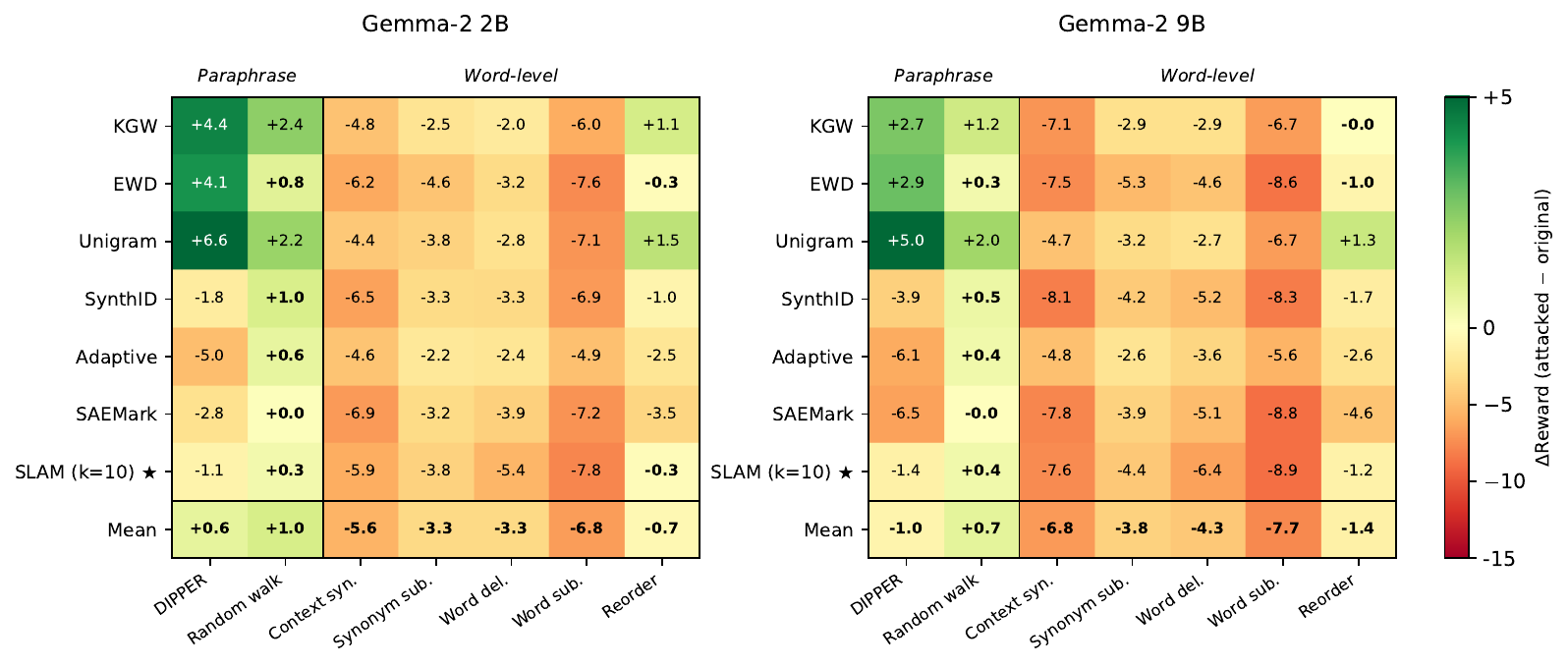}
  \caption{Per-method, per-attack quality change ($\Delta$Reward) on Gemma~2 2B
  (left) and 9B (right). Cells use a diverging colormap: green = quality
  preserved, red = quality destroyed. Attacks are grouped to match
  Figure~\ref{fig:robustness_heatmap} (Paraphrase / Word-level). The bottom
  \emph{Mean} row averages $\Delta$Reward across the seven methods.}
  \label{fig:attack_quality}
\end{figure}

Figure~\ref{fig:attack_quality} shows per-method, per-attack
$\Delta$Reward as a heatmap; attacks whose mean falls below the ${-}3$
utility threshold are flagged as non-practical regardless of post-attack
TPR. Note: averaging hides heavy-tailed degradation. Reorder has
Mean $\Delta\text{Reward} \approx 0$ but $28\%$ of its outputs degrade
by $\geq\!3$ reward points; lucky reorderings cancel harmed cases in the mean.

Table~\ref{tab:attack_validity} renders a verdict per (method, attack)
pair: orange = SLAM evaded but quality drops for SLAM-watermarked text,
yellow = baselines evaded but SLAM robust, red = quality destroyed.
Reading across both panels, DIPPER is the only attack that evades
\slam{} without triggering the quality-destroyed threshold
(cross-method mean $\Delta$Reward $+0.6$ on 2B, $-1.0$ on 9B,
both above $-5$) — but this verdict requires qualification on three fronts.

First, the mean $\Delta$Reward column averages over \emph{all seven}
methods including the baselines. Many baseline watermarks incur
substantial quality degradation, so DIPPER's edits can recover some of
that lost quality and inflate the cross-method average on 2B. When
\slam{}-watermarked outputs are examined in isolation, DIPPER
\emph{lowers} quality by more than one reward point on both 2B and 9B,
so the attack does not preserve quality for the method it purports to
evade.

Second, DIPPER is an 11B parameter model, considerably larger than
the 2B and 9B models being watermarked. All else equal, a stronger
editor paraphrasing weaker-model outputs should improve quality; the
fact that it still degrades \slam{}-watermarked text despite this
size advantage further undermines the claim of quality-preserving
evasion.

Third, the cross-method mean is already negative on 9B ($-1.0$);
accounting for the baseline-inflation effect on 2B (baseline-method
watermarks have DIPPER-recoverable quality penalties that inflate the
average), the picture is uniformly negative once quality-matched
comparisons are used.

Together, these points reframe DIPPER as a partial threat: it can
remove the \slam{} watermark signal (TPR drops to $\approx\!10\text{--}12\%$),
but not without a quality penalty on \slam{}-watermarked text.
Word-level attacks either destroy quality or leave \slam{} robust;
random walk is robust on both models.

\input{table5_attack_validity}

%% file: table5_attack_validity.tex
\begin{table*}[!htbp]
\centering
\caption{Attack validity analysis.
Post-attack TPR (\%) per method (\textbf{bold} = evades, i.e.\ TPR $<80\%$) and mean $\Delta$Reward
\emph{averaged across all seven methods}.
{\color{red}Red} $\Delta$Reward $< -5$ = quality destroyed; attack is invalid regardless of evasion.
Orange = SLAM evaded but quality also drops for SLAM-watermarked text despite positive cross-method mean; yellow = baselines evaded but SLAM robust; white = no valid evasion.}
\label{tab:attack_validity}

\begin{subtable}{\linewidth}
\centering
\caption{Gemma-2 2B}
\resizebox{\linewidth}{!}{%
\begin{tabular}{lrrrrrrrrl}
\toprule
Attack & KGW & EWD & Unigram & SynthID & Adaptive & SAEMark & SLAM & $\Delta$Rew & Verdict \\
\midrule
  DIPPER & \textbf{56} & \textbf{50} & \textbf{51} & \textbf{29} & \textbf{68} & \textbf{4} & \textbf{12} & +0.6 & \cellcolor{orange!20}evaded (SLAM qual.\,$\downarrow$) \\
  Random walk & 98 & 98 & 98 & \textbf{72} & \textbf{77} & \textbf{24} & 100 & +1.1 & \cellcolor{yellow!20}BL only \\
  Reorder & 100 & 100 & 100 & 100 & 97 & \textbf{6} & 100 & -0.7 & \cellcolor{yellow!20}BL only \\
  Synonym sub. & 100 & 100 & 99 & 85 & 100 & \textbf{11} & 100 & -3.3 & \cellcolor{yellow!20}BL only \\
  Word del. & 99 & 99 & 100 & \textbf{50} & 100 & \textbf{2} & 100 & -3.3 & \cellcolor{yellow!20}BL only \\
  \cmidrule{1-10}
  Context syn. & 99 & 98 & 98 & 88 & 100 & \textbf{27} & 100 & \textcolor{red}{-5.6} & \cellcolor{red!10}qual.\ fail \\
  Word sub. & 100 & 98 & 100 & 82 & 100 & \textbf{3} & 100 & \textcolor{red}{-6.8} & \cellcolor{red!10}qual.\ fail \\
  \cmidrule{1-10}
\bottomrule
\end{tabular}}
\end{subtable}

\begin{subtable}{\linewidth}
\centering
\caption{Gemma-2 9B}
\resizebox{\linewidth}{!}{%
\begin{tabular}{lrrrrrrrrl}
\toprule
Attack & KGW & EWD & Unigram & SynthID & Adaptive & SAEMark & SLAM & $\Delta$Rew & Verdict \\
\midrule
  DIPPER & \textbf{64} & \textbf{46} & \textbf{62} & \textbf{14} & \textbf{74} & \textbf{2} & \textbf{10} & -1.0 & \cellcolor{orange!20}evaded (SLAM qual.\,$\downarrow$) \\
  Random walk & 99 & \textbf{5} & 98 & \textbf{68} & 87 & \textbf{10} & 100 & +0.7 & \cellcolor{yellow!20}BL only \\
  Reorder & 100 & \textbf{6} & 99 & 96 & 99 & \textbf{3} & 100 & -1.4 & \cellcolor{yellow!20}BL only \\
  Synonym sub. & 99 & 95 & 98 & \textbf{68} & 100 & \textbf{11} & 100 & -3.8 & \cellcolor{yellow!20}BL only \\
  Word del. & 98 & 95 & 97 & \textbf{40} & 100 & \textbf{8} & 100 & -4.3 & \cellcolor{yellow!20}BL only \\
  \cmidrule{1-10}
  Context syn. & 99 & \textbf{6} & 97 & 81 & 100 & \textbf{10} & 100 & \textcolor{red}{-6.8} & \cellcolor{red!10}qual.\ fail \\
  Word sub. & 99 & 80 & 99 & \textbf{79} & 100 & \textbf{10} & 100 & \textcolor{red}{-7.7} & \cellcolor{red!10}qual.\ fail \\
  \cmidrule{1-10}
\bottomrule
\end{tabular}}
\end{subtable}
\end{table*}

%% file: sections/appendix/03-diversity.tex
\section{Distribution-distance metrics}
\label{app:diversity}

A common concern with watermarking is that biasing the generative
distribution collapses output diversity, producing more uniform
``AI-sounding'' text. We quantify this with three published
distribution-distance measures probing complementary failure modes
(Table~\ref{tab:diversity}).

\input{table_diversity}

\paragraph{Length normalisation.}
All three metrics are computed on full-length generations and
length-normalised by construction (n-gram fractions for distinct-$n$;
mean BLEU-4 for Self-BLEU; cluster-area under the divergence frontier
for MAUVE). This matters because watermark methods produce different
mean lengths under the same prompt set: \slam{}'s outputs are
$\sim\!0.75\times$ baseline length, KGW/EWD/Unigram are
$1.14$--$1.72\times$. Raw $n$-gram counts would conflate watermark
distortion with length differences; the length-normalised metrics
isolate distributional shape.

\paragraph{Why smallest-$|\Delta|$ is the defensible target (distinct-$n$, Self-BLEU).}
The unwatermarked baseline represents the model's natural, optimised
output distribution given the prompt; an imperceptible watermark should
leave that distribution intact. This rules out two failure modes that
an unsigned ``higher is better'' rule would miss: (i) a
\emph{decrease} signals sampling-space restriction (e.g.\ green-list
bias forcing repeated $n$-grams); (ii) an \emph{increase}
typically signals erratic sampling---unnatural synonyms, broken idioms,
or perplexity spikes injected purely to avoid repeating $n$-grams---which we conjecture
is more likely to be text degradation than creativity. Conservatively, the smallest $|\Delta|$ is the
most defensible target on both axes; it certifies the watermark
operates orthogonally to the natural distribution rather than
systematically distorting it.

\paragraph{distinct-$n$ \citep{li2016diversity}.}
The fraction of unique $n$-grams in each watermarked text, averaged
over the prompt set and over $n\!\in\!\{1,\ldots,4\}$. Range $[0,1]$;
intra-text lexical variety. The aggregate hides an $n$-dependent gap:
at $n{=}4$, \slam{} is at $0.98$ on 2B vs.\ KGW $0.42$ and Unigram
$0.24$---the green-list-recurrence signature is much sharper at
higher $n$.

\paragraph{Self-BLEU \citep{zhu2018texygen}.}
Mean BLEU-4 of each watermarked text against the other 99 in the same
set; inter-text redundancy across prompts. A \emph{positive} $\Delta$
flags structural mode collapse (reusing safe high-probability $n$-grams
across prompts); a \emph{negative} $\Delta$ flags erratic per-text
phrasing.

\paragraph{MAUVE \citep{pillutla2021mauve}.}
A distributional-alignment metric: KL-style gap between watermarked and
unwatermarked distributions, computed under GPT-2-large hidden-state
embeddings clustered by $k$-means and reported as the area under the
divergence frontier. Range $[0,1]$; \emph{strictly higher is better}.
MAUVE penalises both Type~I errors (generating text outside the
reference distribution---unnatural watermark forcing) and Type~II
errors (failing to cover the reference---mode collapse). This is the
most direct empirical analogue of SynthID's formal $K$-sequence
non-distortion property \citep{synthid2024}: a perfectly
non-distortionary watermark would yield MAUVE $=1$ in expectation.

\paragraph{Semantic similarity.}
The three metrics above operate at the lexical and distributional surface.
We additionally measure whether watermarking shifts the \emph{meaning} of
completions. For each of 100 evaluation prompts we generate $N{=}10$
unwatermarked completions with temperature sampling ($T{=}1.0$,
$p{=}0.95$) and embed them with Qwen3-Embedding-8B
\citep{qwen3embedding}, a top-tier encoder on the MTEB English
leaderboard \citep{muennighoff2023mteb} (mean score 75.2).
We compute a per-prompt centroid (mean of the $N$ embeddings,
re-normalised) and report the cosine distance from each watermarked
completion to its prompt's centroid.
The \textbf{natural\_variance} row---each of the $N$ unwatermarked
completions measured against its own centroid---is the irreducible
baseline: it captures how much the model's own sampling distribution
varies semantically, independent of any watermark.
A method that matches this baseline causes no detectable semantic drift.

Table~\ref{tab:semantic_sim} shows that \slam{} has the smallest $\Delta$
on both models (+0.035/+0.047 on 2B/9B), while every token-distribution
baseline is at least $+0.054$ from natural\_variance on 9B.
The gap is consistent: the next-closest method on 2B is KGW at $+0.045$,
already $1.3\times$ \slam{}'s drift.
Token-level biasing shifts not only surface form but measurable
meaning; steering the model's internal representation does not.

\paragraph{Pattern across methods.}
\slam{} preserves intra-text variety (distinct-$n$) and overall
distributional shape (MAUVE) while staying within $|\Delta|\!=\!0.017$
of baseline Self-BLEU; no other method clears all three bars.
KGW/EWD/Unigram fail on distinct-$n$ (loss of $\geq\!50\%$ of unique
bigrams) and on MAUVE (near-zero distributional overlap, $0.01$--$0.07$),
the green-list-recurrence signature.
Adaptive is the closest competitor---good on distinct-$n$ and
MAUVE---but is the only method that worsens on grammar
(\S\ref{sec:experiments:main}).
EWD's tight Self-BLEU on 9B ($|\Delta|\!=\!0.010$) is a degenerate
win: its MAUVE is $0.02$, the lowest of any method, so its
watermarked distribution is essentially disjoint from the unwatermarked
one despite per-pair lexical overlap. SynthID's MAUVE 0.20 on 9B is
the strongest of the baselines on this axis but is paired with
$\Delta\text{Reward}={-}2.8$ on 2B and PPL ratio $0.62$.

\input{table_semantic_similarity}

%% file: table_diversity.tex
\begin{table}[t]
\centering
\caption{Distribution-distance metrics: how much each watermark perturbs the unwatermarked output distribution. \textbf{distinct-$n$} \citep{li2016diversity} is mean over $n\!\in\!\{1,...,4\}$ of $\#$unique-$n$-grams$/\#$total-$n$-grams (intra-text lexical variety, length-normalised; range $[0,1]$). \textbf{Self-BLEU} \citep{zhu2018texygen} is mean BLEU-4 of each text against the other 99 in the same set (inter-text redundancy across prompts). \textbf{MAUVE} \citep{pillutla2021mauve} is a distributional-alignment metric: KL-style divergence between watermarked and unwatermarked text distributions under a GPT-2-large embedding (range $[0,1]$; strictly higher = better, since both Type~I errors---unnatural phrasing---and Type~II errors---restricted vocabulary---drive the score down). For distinct-$n$ and Self-BLEU we report the watermarked-set value (WM) and its signed gap to the per-method unwatermarked baseline ($\Delta$), scoring both by smallest $|\Delta|$: an increase signals mode collapse onto shared safe $n$-grams; a decrease signals erratic, high-perplexity sampling. The baselines all share the same canonical text per model (distinct-$n_\text{BL}\!=\!0.901/0.908$ and Self-BLEU$_\text{BL}\!=\!0.072/0.074$ on 2B/9B). \textbf{Bold} = best value in column within the panel; {\small $\bigstar$} = proposed SLAM ($k=10$).}
\label{tab:diversity}
\resizebox{\columnwidth}{!}{%
\begin{tabular}{lrrrrr}
\toprule
 & \multicolumn{2}{c}{\textit{distinct-$n$} (target BL)} & \multicolumn{2}{c}{\textit{Self-BLEU} (target BL)} & \textit{MAUVE} (max) \\
\cmidrule(lr){2-3} \cmidrule(lr){4-5} \cmidrule(lr){6-6}
Method & WM & $\Delta\!\to\!0$ & WM & $\Delta\!\to\!0$ & $\uparrow$ \\
\midrule
\multicolumn{6}{l}{\textit{Gemma-2 2B}}\\
\midrule
  KGW & 0.356 & -0.545 & 0.132 & +0.059 & 0.012 \\
  EWD & 0.322 & -0.579 & 0.107 & +0.035 & 0.017 \\
  Unigram & 0.209 & -0.692 & 0.111 & +0.038 & 0.019 \\
  SynthID & 0.657 & -0.244 & 0.109 & +0.036 & 0.085 \\
  Adaptive & 0.936 & +0.036 & 0.071 & \textbf{-0.002} & 0.130 \\
  SAEMark & 0.644 & -0.257 & 0.111 & +0.038 & \textbf{0.897} \\
  SLAM (k=10) $\bigstar$ & 0.888 & \textbf{-0.013} & 0.058 & -0.015 & 0.194 \\
\midrule
\multicolumn{6}{l}{\textit{Gemma-2 9B}}\\
\midrule
  KGW & 0.420 & -0.488 & 0.125 & +0.051 & 0.065 \\
  EWD & 0.365 & -0.543 & 0.084 & \textbf{+0.010} & 0.021 \\
  Unigram & 0.246 & -0.662 & 0.086 & +0.012 & 0.030 \\
  SynthID & 0.740 & -0.167 & 0.097 & +0.024 & 0.199 \\
  Adaptive & 0.941 & +0.033 & 0.058 & -0.016 & 0.186 \\
  SAEMark & 0.680 & -0.228 & 0.125 & +0.051 & 0.007 \\
  SLAM (k=10) $\bigstar$ & 0.884 & \textbf{-0.024} & 0.057 & -0.017 & \textbf{0.534} \\
\bottomrule
\end{tabular}}
\end{table}

%% file: table_semantic_similarity.tex
\begin{table}[!htbp]
\centering
\caption{Semantic similarity: cosine distance from each watermarked completion to the per-prompt unwatermarked centroid, embedded with Qwen3-Embedding-8B \citep{qwen3embedding} (range $[0,1]$; lower = more semantically similar to the unwatermarked distribution). \textbf{natural\_variance} is the within-prompt spread of 10 independent unwatermarked samples and represents the irreducible baseline: a perfectly non-distortionary watermark would match it. $\Delta$ is the signed gap to this baseline; smallest $|\Delta|$ is the target. \textbf{Bold} = best value in column within the panel; {\small $\bigstar$} = proposed SLAM.}
\label{tab:semantic_sim}
\resizebox{0.7\columnwidth}{!}{%
\begin{tabular}{lrrrr}
\toprule
 & \multicolumn{2}{c}{\textit{Gemma-2 2B}} & \multicolumn{2}{c}{\textit{Gemma-2 9B}} \\
\cmidrule(lr){2-3} \cmidrule(lr){4-5}
Method & cos.\ dist & $\Delta\!\to\!0$ & cos.\ dist & $\Delta\!\to\!0$ \\
\midrule
natural\_variance & 0.316 & --- & 0.312 & --- \\
\midrule
  KGW           & 0.361 & +0.045        & 0.377 & +0.065 \\
  EWD           & 0.373 & +0.057        & 0.375 & +0.064 \\
  Unigram       & 0.377 & +0.061        & 0.366 & +0.054 \\
  SynthID       & 0.378 & +0.062        & 0.371 & +0.060 \\
  Adaptive      & 0.375 & +0.059        & 0.402 & +0.090 \\
  SAEMark       & 0.378 & +0.062        & 0.404 & +0.093 \\
  SLAM (k=10) $\bigstar$ & \textbf{0.350} & \textbf{+0.035} & \textbf{0.359} & \textbf{+0.047} \\
\bottomrule
\end{tabular}}
\end{table}

%% file: sections/appendix/04-ablations.tex
% ─────────────────────────────────────────────────────────────────────────────
\section{Component ablations}
\label{app:ablation}% experiments §4 points here for k×α reference
\label{app:inject_f}% discussion §5 points here for F-sweep reference

Table~\ref{tab:ablation_full} sweeps four orthogonal dimensions of
\slam{} on both PT models, holding every other dimension at its
canonical value ($k{=}10$, $\alpha{=}3/12$ on 2B/9B, $F{=}7$).
$\bigstar$ marks the deployed config per model.

\paragraph{Feature direction.}
Replacing mined SAE decoder directions with re-calibrated random unit
vectors collapses TPR from $100\%$ to $16\%$/$1\%$ on 2B/9B, confirming that
the watermark signal derives from semantic specificity of the mined
features rather than steering magnitude alone.
The near-zero $\Delta\text{Reward}{=}{-0.2}$ on 9B under random directions
is diagnostic: random vectors do not align with any coherent structural
subspace, leaving generation effectively unchanged---and therefore
undetectable.

\paragraph{Quality filter.}
Replacing \texttt{quality\_stats} with uniform weight~1.0 (injecting all
mined features regardless of measured quality impact) preserves 100\% TPR
on both models but degrades quality: $\Delta\text{Reward}$ worsens by
$-1.1$ on 2B ($-1.3\!\to\!-2.4$) and $-1.9$ on 9B ($-1.9\!\to\!-3.8$).
The filter thus earns its keep---eliminating high-cost features at zero
detection cost.

\begin{figure}[!htbp]
  \centering
  \includegraphics[width=\linewidth]{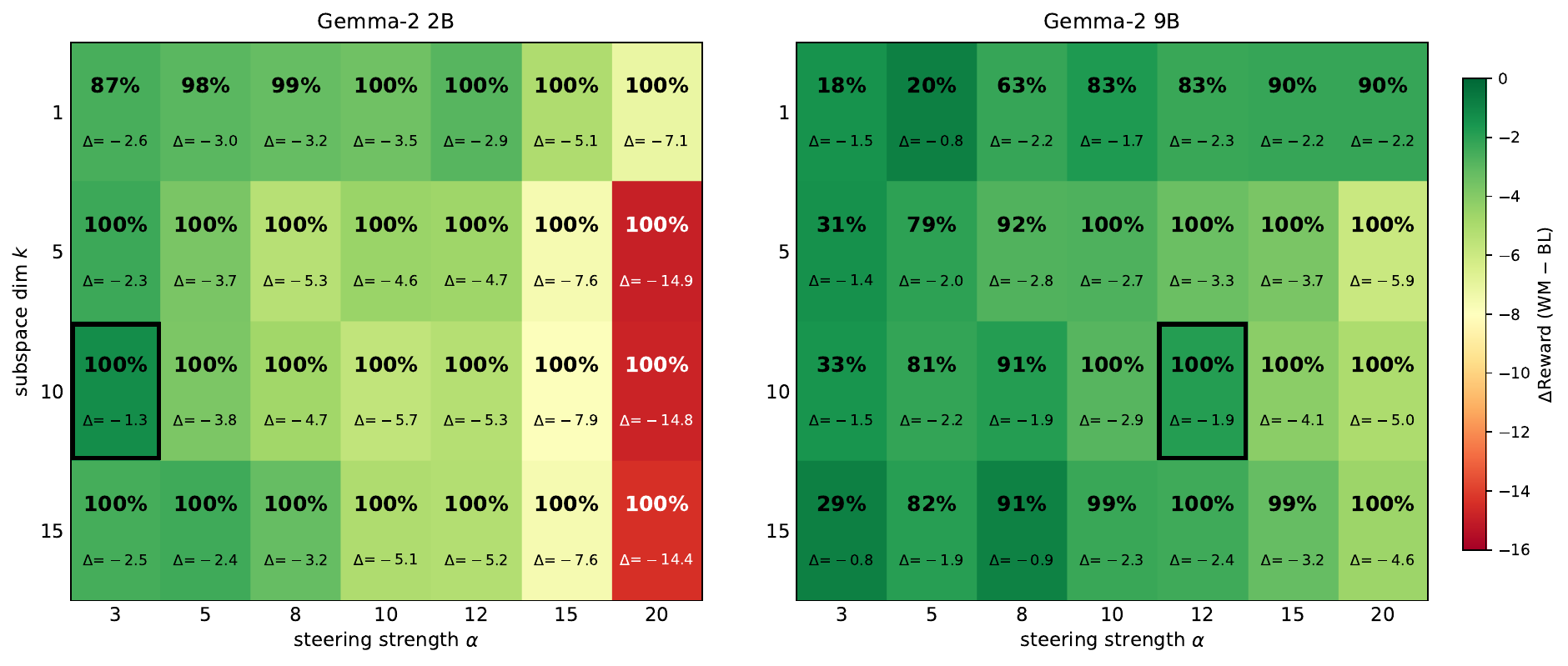}
  \caption{$k \times \alpha$ sweep on Gemma~2 2B (left) and 9B (right).
  Each cell reports detection rate (TPR \%, large) and quality gap
  ($\Delta$Reward, small). Color encodes $\Delta$Reward (greener = less
  quality cost). Bold-bordered cell: proposed config per model
  ($k{=}10$, $\alpha{=}3$ on 2B; $k{=}10$, $\alpha{=}12$ on 9B).}
  \label{fig:k_alpha}
\end{figure}

\paragraph{Subspace dimension $k$ and the $k\!\times\!\alpha$ frontier.}
Figure~\ref{fig:k_alpha} shows the full $k\!\times\!\alpha$ grid.
$k{=}10$ is optimal on both models: it achieves 100\% TPR at the
lowest $\alpha$ (hence lowest quality cost) on 2B ($\alpha{=}3$,
$\Delta\text{Reward}{=}{-1.3}$), and achieves the best quality at 100\% TPR
on 9B ($\alpha{=}12$, $\Delta\text{Reward}{=}{-1.9}$; $k{=}5$ also
reaches 100\% TPR but at $\alpha{=}10$ with $\Delta\text{Reward}{=}{-2.7}$).
At 9B, $k{=}1$ caps at 90\% TPR even at $\alpha{=}20$, consistent with
structural concepts being distributed across many weakly-discriminative
SAE features at larger scale; composing $k$ orthogonal directions is
what recovers full detection.

\begin{table*}[!htbp]
\centering
\caption{Component ablations for Gemma-2 2B and 9B (PT).
Each panel varies one dimension; all others fixed at the canonical config
($k{=}10$, $\alpha{=}3$ on 2B / $\alpha{=}12$ on 9B, $F{=}7$).
$\bigstar$ = canonical \slam{} setting for that model.
TPR (\%) at $\approx\!2.3\%$ FPR (cal-$z > 2.0$). \textbf{Bold} = $\geq\!99\%$.
For the subspace dim.\ $k$ panel, each row shows the best-quality $\alpha$
at TPR\,$\geq\!90\%$; the full $k\!\times\!\alpha$ grid is in
Figure~\ref{fig:k_alpha}.}
\label{tab:ablation_full}
\small
\begin{tabular}{llrrrr}
\toprule
Ablation & Variant
  & \multicolumn{2}{c}{Gemma-2 2B}
  & \multicolumn{2}{c}{Gemma-2 9B} \\
\cmidrule(lr){3-4}\cmidrule(lr){5-6}
 & & TPR (\%) & $\Delta$Rew & TPR (\%) & $\Delta$Rew \\
\midrule
%% ── Feature direction ─────────────────────────────────────────────
\multirow{2}{*}{Feature direction}
  & Mined SAE directions ($\bigstar$) & \textbf{100} & $-1.3$ & \textbf{100} & $-1.9$ \\
  & Random unit vectors               &           16 & $-1.5$ &            1 & $-0.2$ \\
\midrule
%% ── Quality filter ────────────────────────────────────────────────
\multirow{2}{*}{Quality filter}
  & With quality filter ($\bigstar$) & \textbf{100} & $-1.3$ & \textbf{100} & $-1.9$ \\
  & Without quality filter           & \textbf{100} & $-2.4$ & \textbf{100} & $-3.8$ \\
\midrule
%% ── Subspace dimension k ──────────────────────────────────────────
\multirow{4}{*}{Subspace dim.\ $k$}
  & $k{=}1$,  $\alpha{=}12/15$            & \textbf{100} & $-2.9$ &           90 & $-2.2$ \\
  & $k{=}5$,  $\alpha{=}3/10$            & \textbf{100} & $-2.3$ & \textbf{100} & $-2.7$ \\
  & $k{=}10$, $\alpha{=}3/12$ ($\bigstar$)& \textbf{100} & $-1.3$ & \textbf{100} & $-1.9$ \\
  & $k{=}15$, $\alpha{=}5/8$             & \textbf{100} & $-2.4$ &           91 & $-0.9$ \\
\midrule
%% ── Steering strength α ───────────────────────────────────────────
\multirow{7}{*}{Steering strength $\alpha$ ($k{=}10$)}
  & $\alpha{=}3$\phantom{0}  ($\bigstar_{\mathrm{2B}}$) & \textbf{100} & $-1.3$ &           33 & $-1.5$ \\
  & $\alpha{=}5$\phantom{0}                             & \textbf{100} & $-3.8$ &           81 & $-2.2$ \\
  & $\alpha{=}8$\phantom{0}                             & \textbf{100} & $-4.7$ &           91 & $-1.9$ \\
  & $\alpha{=}10$                                       & \textbf{100} & $-5.7$ & \textbf{100} & $-2.9$ \\
  & $\alpha{=}12$ ($\bigstar_{\mathrm{9B}}$)            & \textbf{100} & $-5.3$ & \textbf{100} & $-1.9$ \\
  & $\alpha{=}15$                                       & \textbf{100} & $-7.9$ & \textbf{100} & $-4.1$ \\
  & $\alpha{=}20$                                       & \textbf{100} & $-14.8$& \textbf{100} & $-5.0$ \\
\midrule
%% ── Features injected per document F ─────────────────────────────
\multirow{4}{*}{Features per doc.\ $F$ ($k{=}10$)}
  & $F{=}7$\phantom{0}  ($\bigstar$) & \textbf{100} & $-1.3$ & \textbf{100} & $-1.9$ \\
  & $F{=}10$                         & \textbf{100} & $-2.2$ & \textbf{100} & $-3.4$ \\
  & $F{=}15$                         & \textbf{100} & $-3.4$ & \textbf{100} & $-4.6$ \\
  & $F{=}20$                         & \textbf{100} & $-3.0$ & \textbf{100} & $-2.7$ \\
\bottomrule
\end{tabular}
\end{table*}

\paragraph{Steering strength $\alpha$.}
On 2B, any $\alpha\!\geq\!3$ with $k{=}10$ achieves 100\% TPR;
$\alpha{=}3$ minimises quality cost ($\Delta\text{Reward}{=}{-1.3}$).
On 9B, $\alpha{=}10$ is the minimum for 100\% TPR; $\alpha{=}12$
improves quality to $-1.9$ without losing detection.
Very high $\alpha$ is harmful: $\alpha{=}20$ on 2B degrades
$\Delta\text{Reward}$ to $-14.8$, worse than any baseline in Table~1.

\paragraph{Features injected per document $F$.}
Clean TPR saturates at $F{=}7$ on both models, so larger $F$ buys no
detection headroom.
$\Delta$Reward strictly worsens through $F{=}15$ on both models
($-1.3\!\to\!-3.4$ on 2B; $-1.9\!\to\!-4.6$ on 9B), with a slight
bounce-back at $F{=}20$ as the L2-normalised composite steering vector
smooths out.
On 9B, larger $F$ improves paraphrase survival (DIPPER TPR
$10\%\!\to\!57\%$ at $F{=}15$); on 2B, post-attack TPR is flat-to-decreasing.
We retain $F{=}7$ as the canonical setting; $F{=}15$ is a configurable
knob for practitioners who weigh paraphrase robustness over clean quality.

%% file: sections/appendix/05-timing.tex
\section{Computational overhead}
\label{app:timing}

Table~\ref{tab:timing} reports generation and detection time per text for
each method on Gemma~2 PT models, measured in isolation on a single A100-40GB GPU
with \texttt{max\_new\_tokens=200}, 3 warm-up texts, and 20 timed texts per method.

\slam{}'s end-to-end overhead is modest: generation is
$9.3$\,s vs.\ the unwatermarked baseline's $5.4$\,s on 2B ($1.7\times$) and
$14.3$\,s vs.\ $8.7$\,s on 9B ($1.6\times$), the slowdown coming from the
activation-steering hook. Detection runs one full model forward pass
over the candidate text (no autoregressive sampling) followed by a
cheap $O(d)$ projection of the residual onto the bank direction; the
forward pass dominates ($\sim\!420$\,ms on 2B, $\sim\!640$\,ms on 9B)
and amortises the same infrastructure loaded for generation.
KGW, Unigram, and SynthID detect via a token-ID statistical scan
(no model needed); EWD and Adaptive, like \slam{}, require a model
forward pass. The Adaptive outlier: its detection routine takes
$\sim\!10.6$\,s per text ($25\times$ \slam{}'s detection cost on 2B,
$17\times$ on 9B),
making it impractical for high-throughput deployment regardless of its
robustness profile.

\input{table_timing}

%% file: table_timing.tex
\begin{table}[!htbp]
  \centering
  \caption{Computational overhead per text (PT models). Each method benchmarked in isolation; GPU memory cleared between methods.}
  \label{tab:timing}
  \begin{threeparttable}
  \resizebox{\columnwidth}{!}{%
  \begin{tabular}{llrrlll}
    \toprule
    Model & Method & Gen.\,time\,(s)\tnote{a} & Tok/s & Det.\,req. & Det.\,time\tnote{b} & Gen.\,ovhd.\tnote{c} \\
    \midrule
    Gemma-2 2B & Unwatermarked & 5.4$\pm$0.4 & 37 & --- & --- & 1.0$\times$ \\
     & SLAM $\bigstar$ & 9.3$\pm$1.6 & 21 & GPU fwd. & 418$\pm$33\,ms & 1.7$\times$ \\
     & KGW & 5.6$\pm$0.0 & 36 & token scan & 51$\pm$0\,ms & 1.0$\times$ \\
     & EWD & 5.4$\pm$0.7 & 35 & GPU fwd. & 77$\pm$5\,ms & 1.0$\times$ \\
     & Unigram & 5.5$\pm$0.4 & 36 & token scan & 4$\pm$0\,ms & 1.0$\times$ \\
     & SynthID & 5.5$\pm$1.1 & 32 & token scan & 38$\pm$7\,ms & 1.0$\times$ \\
     & Adaptive & 20.8$\pm$1.3 & 10 & GPU fwd. & 10.6$\pm$0.8\,s & 3.9$\times$ \\
     & SAEMark & 120.3$\pm$73.5 & 2 & GPU fwd. & 7.0$\pm$0.4\,s & 22.4$\times$ \\
    \midrule
    Gemma-2 9B & Unwatermarked & 8.7$\pm$0.0 & 23 & --- & --- & 1.0$\times$ \\
     & SLAM $\bigstar$ & 14.3$\pm$1.5 & 14 & GPU fwd. & 641$\pm$7\,ms & 1.6$\times$ \\
     & KGW & 8.9$\pm$0.1 & 22 & token scan & 50$\pm$0\,ms & 1.0$\times$ \\
     & EWD & 9.0$\pm$0.3 & 22 & GPU fwd. & 100$\pm$2\,ms & 1.0$\times$ \\
     & Unigram & 9.0$\pm$0.1 & 22 & token scan & 4$\pm$0\,ms & 1.0$\times$ \\
     & SynthID & 8.8$\pm$1.2 & 21 & token scan & 38$\pm$5\,ms & 1.0$\times$ \\
     & Adaptive & 25.0$\pm$1.5 & 9 & GPU fwd. & 10.8$\pm$1.0\,s & 2.9$\times$ \\
     & SAEMark & 171.9$\pm$120.6 & 2 & GPU fwd. & 7.2$\pm$0.2\,s & 19.8$\times$ \\
    \bottomrule
  \end{tabular}}
  \begin{tablenotes}\footnotesize
    \item[a] Mean$\pm$std over 20 texts (3 warm-up), \texttt{max\_new\_tokens=200}, single A100-40GB, no concurrent jobs.
    \item[b] Detection-only time after generation; SLAM uses one LLM forward pass.
    \item[c] Generation time relative to the unwatermarked baseline (greedy decoding, no hooks).
  \end{tablenotes}
  \end{threeparttable}
\end{table}

%% file: sections/appendix/06-implementation_details.tex
\section{Implementation details}
\label{app:impl}

\subsection{Mining configuration}
\label{app:impl:mining}

\paragraph{Source pool and per-phenomenon cap.}
The verified contrastive bank contains $\sim$88{,}362 pairs across the 104
phenomena (after template generation, cross-domain expansion, and automatic
verification). To prevent high-volume phenomena (e.g.\ morphological
suffixes) from dominating composite-score statistics over scarcer
ones, the mining stage caps each phenomenon at \textbf{200 pairs} (sampled
uniformly without replacement, fixed seed $42$); phenomena with fewer than
200 verified pairs are used in full. The post-cap mined dataset is
46{,}579 pairs.

\paragraph{Filter thresholds.}
Two filters are applied in series. The \textbf{gap-consistency} filter
retains features whose causal gap is positive on $\geq\!80\%$ of test
prompts. The \textbf{composite} filter then retains features whose composite score
$s_j = |\Delta\mu_j| \times \text{purity}_j \times \text{consistency}_j$
exceeds $0.05$. Subspace mining uses PCA with $k{=}15$ bidirectional
components; the $k$ used for steering is selected post-hoc from the
$k\!\in\!\{1,5,10,15\}$ ablation.

\paragraph{Mining funnel counts.}
On Gemma~2 2B, 51{,}078 candidate SAE features (summed across 13 layers and
104 phenomena) survive the gap-consistency pre-filter at 9.6\% (4{,}922),
of which 42.7\% (2{,}104) clear the composite threshold---roughly 4\% of
raw candidates overall.
9B is comparable: 50{,}819 $\to$ 4{,}283 $\to$ 1{,}057 (8.4\% then 24.7\%).

\paragraph{Final bank sizes (post-mining, pre-calibration).}
Per-model final bank, in phenomena/composite-direction count:
2B = 46 phenomena; 9B = 60; 2B-IT = 73--75 (per-$k$); 9B-IT = $\sim\!76$.
Bidirectional mining doubles the per-phenomenon direction count
(forward and reverse polarity), so the steerable per-document pool
is roughly twice the phenomenon count.

\subsection{Contrastive phenomena (104)}
\label{app:impl:phenomena}

The full set of 104 phenomena mined for the production banks
(90 imported from LinguaLens \citep{jing2025lingualens},
14 hand-authored) is grouped below by linguistic level.

\begin{description}\setlength{\itemsep}{0pt}
\item[Morphology — suffixes/prefixes:]
\emph{adjectival}, \emph{adverbial}, \emph{agentive}, \emph{nominal},
and \emph{verbal} suffixes;
\emph{negation}, \emph{temporal}, \emph{quantitative},
\emph{spatial-or-directional}, and \emph{degree} prefixes.
\item[Syntactic alternations:]
\emph{voice}, \emph{PP-fronting}, \emph{particle placement},
\emph{adverb placement}, \emph{extraposition}, \emph{dative alternation},
\emph{heavy-NP shift}, \emph{cleft constructions},
\emph{subject--auxiliary inversion}, \emph{subject--verb inversion},
\emph{locative inversion}, \emph{do-support}, \emph{split infinitives},
\emph{object expletives}, \emph{expletive}, \emph{coordination},
\emph{relative clauses}, \emph{non-defining relative clauses},
\emph{noun clauses}, \emph{clausal subjects}, \emph{appositives}.
\item[Tense--aspect--mood:]
past forms (\emph{past}, \emph{past-tense}, \emph{past-perfect},
\emph{past-participle}); present forms (\emph{present-perfect},
\emph{present-participle}, \emph{progressive aspect});
future forms (\emph{future}, \emph{future-perfect},
\emph{future-progressive}, \emph{futurates});
aspectual contrasts (\emph{punctual} vs.\ \emph{durative},
\emph{telic} vs.\ \emph{atelic}, \emph{static} vs.\ \emph{dynamic});
mood (\emph{first conditional}, \emph{subjunctive}, \emph{imperative},
\emph{optative}, \emph{epistemic} and \emph{deontic} modality).
\item[Speech acts:]
\emph{commissive}, \emph{declaration}, \emph{directive},
\emph{expressive}, \emph{representative}, \emph{indirect speech},
\emph{politeness}.
\item[Agreement and structure:]
\emph{subject--verb agreement}, \emph{determiner--noun agreement},
\emph{argument structure},
\emph{transitive/intransitive/middle/linking verb}, \emph{copular be},
\emph{direct object}, \emph{resultative}, \emph{possessive form},
\emph{s-/of-genitive}, \emph{nominal adverbials}, \emph{count/mass nouns},
\emph{universal/existential quantifiers}, \emph{quantifier},
\emph{comparative}, \emph{superlative}, \emph{negation},
\emph{existential}, \emph{person}, \emph{spatial}, \emph{temporal}.
\item[Discourse and pragmatics:]
\emph{discourse markers}, \emph{transitional}, \emph{intensifiers},
\emph{emphatic structure}, \emph{elliptical sentences},
\emph{echo/tag/interrogative questions}, \emph{turn-taking},
\emph{anaphor}, \emph{deixis}, \emph{referring}, \emph{given/known information}.
\item[Figurative:]
\emph{metaphor}, \emph{personification}, \emph{hyperbole},
\emph{euphemism}, \emph{synecdoche}, \emph{non-synecdoche metonymy}.
\item[Other:]
\emph{factives}, \emph{irregular forms}.
\end{description}

\begin{figure}[!htbp]
  \centering
  \includegraphics[width=\linewidth]{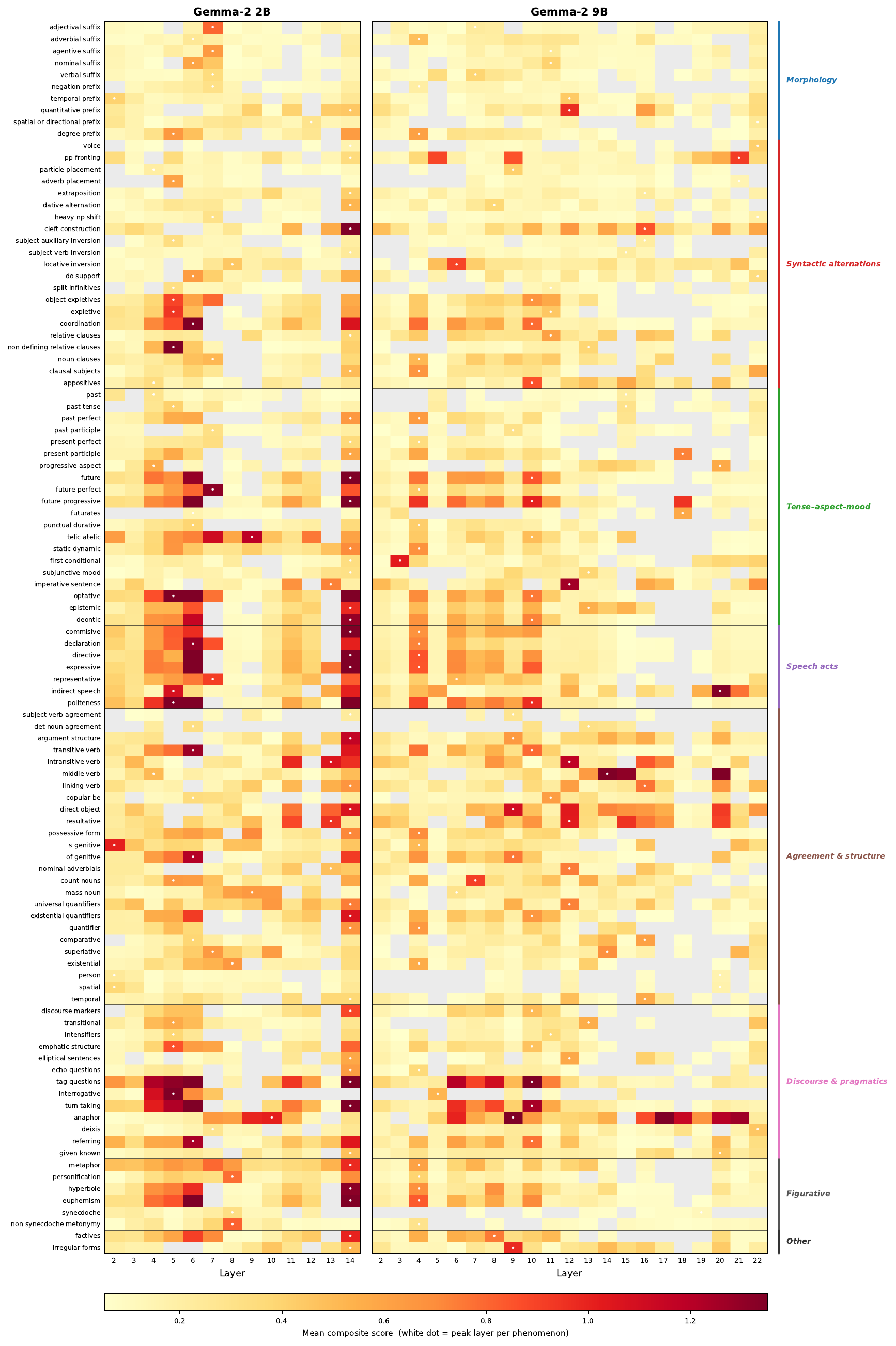}
  \caption{Mean composite score (contrastive $\times$ purity $\times$ consistency) per
  phenomenon $\times$ layer for Gemma-2 2B (left) and 9B (right).
  White dots mark the layer with the highest mean composite score for each phenomenon.
  Grey cells indicate no features survived the composite threshold at that layer.
  Phenomena are grouped by linguistic level; boundaries marked with horizontal lines.}
  \label{fig:phenomenon_layers}
\end{figure}

Figure~\ref{fig:phenomenon_layers} shows the mean composite score per phenomenon $\times$ layer
for both models; the white dot marks the peak layer for each phenomenon.
Morphological phenomena (top block) concentrate in early layers ($\leq\!7$ on 2B,
$\leq\!8$ on 9B), while syntactic alternations and tense--aspect--mood phenomena tend
to peak at later layers (13--14 on 2B; spread over 9--22 on 9B), consistent with
the established layer-specificity of syntactic representations in transformer residual streams.

\subsection{SAE selection}
\label{app:impl:sae}

We use Gemma~Scope residual-stream SAEs (post-MLP residual stream) with width 16k and per-layer
$L_0 \in [53, 83]$ on 2B and $L_0 \in [37, 75]$ on 9B.
``Canonical'' Gemma~Scope checkpoints are avoided where their $L_0 > 100$
(layers 4, 6, 8, 10, 12, 22 on 9B); lower-$L_0$ per-layer checkpoints are
used instead.
For Gemma~2 2B we mine layers $2$--$14$; for 9B, layers $2$--$22$.

\subsection{Steering layer}
\label{app:impl:steer}

Each selected feature carries a layer attribute, and the
injector registers one hook per distinct layer among the $K$ selected features,
summing per-layer contributions before applying the $\alpha$-scaled update.
Injection is therefore \emph{multi-layer}: for a given document the $K{=}7$
HMAC-selected features may target several different layers simultaneously.
Figure~\ref{fig:phenomenon_layers} and the peak-layer summary above show that
morphological phenomena tend to peak in early layers ($\leq\!7$ on 2B) while
syntactic alternations and tense--aspect--mood phenomena peak later (layers
13--14 on 2B; 9--22 on 9B), so the effective injection depth is
phenomenon-dependent and spreads across many residual-stream layers.

\subsection{HMAC-keyed feature selection}
\label{app:impl:hmac}

Per-document feature selection is keyed by
$\text{HMAC-SHA256}(\text{key},\, \text{doc\_id} \mathbin{\|} \text{sentence\_idx})$,
where the sentence index allows the detector to recover per-sentence
feature assignments under sentence-deletion or reorder attacks.
The document identifier is a unique-per-output string (UUID, request hash,
or model-side counter); in our evaluation it is set to a fixed per-prompt
string. The secret key is fixed per experiment; rotating it produces
non-overlapping watermark assignments.

\subsection{Eval-time constants}
\label{app:impl:eval}

The evaluation pipeline uses the following fixed constants:
$K\!=\!7$ features selected per document (HMAC-strength-proportional
sampling over a pool of size 10 with a top-tier anchor of size 5,
where the top-tier anchor is the 5 highest-composite-score features
that are always eligible for selection, and the remaining pool slots
draw from the next 5 by score); selection temperature 0.3; per-feature
pre-filter threshold $z_j \geq 0.5$; detection threshold
$\calz \geq 2.0$. Generation uses temperature 0.7, top-$p$ 0.9.

\paragraph{Early-stop and degeneracy filter.}
The generation loop samples up to $N{=}4$ candidates and returns the first
whose calibrated score clears $\calz \geq 2.0$.
If no candidate clears after $N$ attempts, the candidate with the highest
$\calz$ is returned as a deterministic fallback (always the same text for a
given prompt and key).
The degeneracy filter rejects any candidate whose generated continuation is
empty, shorter than a minimum token length, or a near-verbatim repetition of
the prompt prefix; rejected candidates consume one slot of the $N{=}4$ budget
and trigger an additional generation attempt.

\subsection{Calibration corpus}
\label{app:impl:calib}

Per-feature null statistics
$(\mu_j^{\text{null}}, \sigma_j^{\text{null}})$ are estimated on the
unwatermarked baseline-text set (the unwatermarked Gemma-2 model run on
the same prompt set used in evaluation).
The bank-level null distribution
$(\mu^{\text{null}}_{\text{raw}}, \sigma^{\text{null}}_{\text{raw}})$ is
estimated by running the same $K\!=\!7$ HMAC selection over $N\!=\!100$
baseline texts and computing the aggregated $z$-score, matching
the eval-time selection regime exactly. The 2.3\% FPR figure reported
in the main text is the analytic Gaussian upper-tail at $\calz\!=\!2.0$
under this calibrated null; empirical FPR estimated on the same 100
baseline texts is consistent within the $\pm 1\%$ confidence interval
expected at $N\!=\!100$.

\subsection{Conditional PPL ratio}
\label{app:impl:ppl}

Table~\ref{tab:main_results}'s PPL ratio is conditional perplexity on
the generated continuation tokens only, with the shared prompt prefix
masked so that prompt tokens do not dilute the watermark-induced shift. The scoring model is the same
generative LM that produced the text (Gemma-2 2B or 9B in eval mode),
not a separate reference LM such as GPT-2.
This choice has a precise interpretation: the PPL ratio measures how
much the watermark distorts \emph{the same model's} natural next-token
distribution, conditioned on its own prompt; ratio $= 1$ means the
watermark is invisible to the LM that produced it,
${<}\!1$ indicates repetitive text the LM finds easier to predict
than baseline, and ${>}\!1$ indicates text the LM finds harder
to predict.
Using a fixed-reference scorer like GPT-2 instead would conflate
watermark distortion with the systematic distribution gap between
GPT-2 and the actual generative model.

\subsection{Grammar-error baseline}
\label{app:impl:grammar}

Table~\ref{tab:main_results}'s Gram.\ Err.\ column reports
LanguageTool (en-US) errors per 100 tokens on watermarked
text (length-normalised). The unwatermarked baseline rate is
$1.39$ on 2B PT, $1.31$ on 9B PT, $0.55$ on 2B-IT, $0.49$ on 9B-IT.

\subsection{Baseline watermark hyperparameters}
\label{app:impl:baselines}

All five token-distribution baselines use the default configurations
shipped with MarkLLM v0.1.5 \citep{markllm}, applied as-is to
both the 2B and 9B models without per-model tuning.
Key generation parameters:

\begin{itemize}\setlength{\itemsep}{1pt}\setlength{\parsep}{0pt}
  \item \textbf{KGW} \citep{kirchenbauer2023watermark}:
        $\gamma{=}0.5$ (green-list fraction),
        $\delta{=}2.0$ (logit bias),
        prefix length $1$, time-based seeding, left-window context.
  \item \textbf{EWD} \citep{kuditipudi2023robust}:
        $\gamma{=}0.5$, $\delta{=}2.0$, prefix length $1$
        (MarkLLM's EWD uses the same green-list-bias scheme as KGW
        rather than the distortion-free exponential minimum sampling of
        the original; see \S\ref{sec:experiments:setup}).
  \item \textbf{Unigram} \citep{zhao2023provable}:
        $\gamma{=}0.5$, $\delta{=}2.0$, fixed global green list.
  \item \textbf{SynthID} \citep{synthid2024}:
        $n$-gram context length $5$, $30$ independent hash keys,
        sampling table size $65536$,
        non-distortionary tournament sampling with $2$ leaves
        (equivalent to the $K{=}2$ binary tournament), mean detector,
        context history size $1024$.
  \item \textbf{Adaptive} \citep{liu2024adaptive}:
        $\alpha{=}2.0$, $\delta_0{=}1.0$, $\delta{=}1.5$,
        entropy measure threshold $50$,
        repetition penalty $1.1$.
\end{itemize}

SAEMark is run with its own published defaults.
All detection uses our calibrated threshold ($\calz \geq 2.0$,
Appendix~\ref{app:impl:calib}) rather than the MarkLLM native
$z$-threshold of $4.0$, ensuring a consistent $\approx\!2.3\%$ FPR
comparison across all methods.

\subsection{Metric uncertainty ($N{=}100$)}
\label{app:impl:sem}

All metrics in Table~\ref{tab:main_results} are computed over $N{=}100$
texts per condition.
Standard errors (SE) across these texts are consistent in magnitude across
methods and models:
Cal-Z SE $= 0.04$--$0.48$ (calibrated z-score, higher end driven by
SAEMark 9B's collapsed detector);
$\Delta$Reward SE $= 0.64$--$0.92$ reward points (paired per-text
difference);
PPL ratio SE $= 0.01$--$0.06$.
TPR uses a binomial estimator
($\text{SE} = \sqrt{p(1-p)/N}$), which is $\leq 0.01$ for all methods at
$99$--$100\%$ TPR and at most $0.03$ for SAEMark 9B (13\% TPR).
Grammar error rate is length-normalised per 100 tokens; per-text variance
is not separately reported.
All cross-method $\Delta$Reward gaps exceeding $\approx\!2$ reward points
are statistically reliable at the 95\% level; the tightest comparison
(\slam{} vs.\ SynthID on 9B, gap $\approx\!1.7$ points, $z \approx\!1.9$)
is borderline and should be interpreted with appropriate caution.

\subsection{Reproducibility}
\label{app:impl:repro}

HMAC seeds are SHA-256-based; the secret key is fixed per experiment.
All generation uses
temperature 0.7, top-$p$ 0.9, max-new-tokens 200; mining and
calibration sampling use seed $42$.
Code and feature banks will be released upon publication.
Diversity metrics in Appendix~\ref{app:diversity} use standard
implementations: distinct-$n$ on whitespace-split tokens; Self-BLEU via
NLTK with smoothing method 1 over $n\!\in\!\{1,...,4\}$; MAUVE via the
official MAUVE package with the GPT-2-large featurizer.

%% file: sections/appendix/07-attack_details.tex
\section{Attack details}
\label{app:attacks}

All attacks use a fixed random seed (42) for reproducibility.
Robustness is reported on 100 prompts per condition.

\paragraph{Paraphrase attacks.}
\textbf{DIPPER} \citep{krishna2024paraphrasing}:
\texttt{kalpeshk2011/dipper-paraphraser-xxl} (T5-XXL backbone),
lexical-diversity 60, order-diversity 0; the strongest neural paraphraser
in our suite, restructures both lexical choice and sentence order while
preserving meaning.
\textbf{Random walk} \citep{zhang2023watermarks}: 50 iterations of T5-XL span-masking
(\texttt{google/t5-v1\_1-xl}, span length 6 words; one randomly-positioned
span per iteration; nucleus sampling top-$p\!=\!0.95$ for the infill)
with a Skywork-Reward-V2  per-step acceptance
oracle: a candidate is accepted if its reward score stays within
0.5 reward points (Skywork-Reward-V2 scale) of the \emph{initial}
(un-attacked) text---fixed reference, not running---to prevent unbounded
quality drift over steps.

\paragraph{Word-level attacks.}
All word-level attacks were used with default configurations set in MarkLLM 
\citep{markllm}, an open-source toolkit for LLM watermarking.
\textbf{Synonym substitution} (WordNet): each word has a 30\% chance of
being replaced by a random WordNet lemma; if no synonym exists, the
original is kept, so the \emph{effective} substitution rate is materially
lower than 30\%.
\textbf{Context synonym substitution} (BERT-MLM): first restricts to
content words ($\geq 4$ characters, alphabetic), then sequentially masks
30\% of those positions and fills each from \texttt{bert-base-uncased}'s
top-1 logit, \emph{forcing} a different token (identity replacements
rejected). This makes context-synonym strictly more aggressive than plain
synonym substitution despite the ``contextual'' label: it hits the full
30\% of dense content words and BERT's top-1 prediction tends to favor
high-frequency neighbors that are locally fluent but semantically vaguer
than the original---which Skywork-Reward-V2 penalizes more heavily than
the awkward-but-rare WordNet swaps. This is why context\_synonym shows a
\emph{worse} $\Delta$Reward than synonym\_sub in
Figure~\ref{fig:attack_quality}.
\textbf{Word substitution}: target rate 15\%, replacement drawn from
the per-text-set vocabulary (alphabetic tokens of length $>\!3$ aggregated
across the 100 watermarked texts under attack), filtered to $\pm 2$
characters of the original word's length; if no length-match exists in
vocab the original is kept, so the effective rate is slightly below 15\%.
This introduces semantically incoherent tokens, which is why it dominates
the bottom of the quality ranking.
\textbf{Word deletion}: each word independently dropped with probability
0.3.
\textbf{Sentence reorder}: NLTK sentence-tokenize, then random shuffle of
sentences within each text. Affects discourse coherence but preserves
local token statistics; most token-distribution watermarks survive
(KGW/Unigram at $99$--$100\%$ TPR, Table~\ref{tab:robustness}).

%% file: NEURIPS2026/checklist.tex
\section*{NeurIPS Paper Checklist}

\begin{enumerate}

\item {\bf Claims}
    \item[] Question: Do the main claims made in the abstract and introduction accurately reflect the paper's contributions and scope?
    \item[] Answer: \answerYes{}
    \item[] Justification: Each stated contribution and headline number (100\% TPR, $\Delta$Reward $=-1.3/-1.9$, DIPPER vulnerability with quality cost) maps directly to a result in \S\ref{sec:experiments} and Table~\ref{tab:main_results}.
    \item[] Guidelines:
    \begin{itemize}
        \item The answer \answerNA{} means that the abstract and introduction do not include the claims made in the paper.
        \item The abstract and/or introduction should clearly state the claims made, including the contributions made in the paper and important assumptions and limitations. A \answerNo{} or \answerNA{} answer to this question will not be perceived well by the reviewers.
        \item The claims made should match theoretical and experimental results, and reflect how much the results can be expected to generalize to other settings.
        \item It is fine to include aspirational goals as motivation as long as it is clear that these goals are not attained by the paper.
    \end{itemize}

\item {\bf Limitations}
    \item[] Question: Does the paper discuss the limitations of the work performed by the authors?
    \item[] Answer: \answerYes{}
    \item[] Justification: \S\ref{sec:discussion} (Limitations) covers white-box access requirement, DIPPER vulnerability (including the quality cost caveat and DIPPER's 11B size advantage), single-bit encoding, and the ${\sim}{-4}$ $\Delta$Reward penalty when applying PT SAEs to IT models.
    \item[] Guidelines:
    \begin{itemize}
        \item The answer \answerNA{} means that the paper has no limitation while the answer \answerNo{} means that the paper has limitations, but those are not discussed in the paper.
        \item The authors are encouraged to create a separate ``Limitations'' section in their paper.
        \item The paper should point out any strong assumptions and how robust the results are to violations of these assumptions (e.g., independence assumptions, noiseless settings, model well-specification, asymptotic approximations only holding locally). The authors should reflect on how these assumptions might be violated in practice and what the implications would be.
        \item The authors should reflect on the scope of the claims made, e.g., if the approach was only tested on a few datasets or with a few runs. In general, empirical results often depend on implicit assumptions, which should be articulated.
        \item The authors should reflect on the factors that influence the performance of the approach. For example, a facial recognition algorithm may perform poorly when image resolution is low or images are taken in low lighting. Or a speech-to-text system might not be used reliably to provide closed captions for online lectures because it fails to handle technical jargon.
        \item The authors should discuss the computational efficiency of the proposed algorithms and how they scale with dataset size.
        \item If applicable, the authors should discuss possible limitations of their approach to address problems of privacy and fairness.
        \item While the authors might fear that complete honesty about limitations might be used by reviewers as grounds for rejection, a worse outcome might be that reviewers discover limitations that aren't acknowledged in the paper. The authors should use their best judgment and recognize that individual actions in favor of transparency play an important role in developing norms that preserve the integrity of the community. Reviewers will be specifically instructed to not penalize honesty concerning limitations.
    \end{itemize}

\item {\bf Theory assumptions and proofs}
    \item[] Question: For each theoretical result, does the paper provide the full set of assumptions and a complete (and correct) proof?
    \item[] Answer: \answerNA{}
    \item[] Justification: Empirical paper; the detection statistic (Stouffer $z$-score, Eqs.~\ref{eq:zscore}--\ref{eq:calz}) is a standard construction validated empirically via calibrated FPR.
    \item[] Guidelines:
    \begin{itemize}
        \item The answer \answerNA{} means that the paper does not include theoretical results.
        \item All the theorems, formulas, and proofs in the paper should be numbered and cross-referenced.
        \item All assumptions should be clearly stated or referenced in the statement of any theorems.
        \item The proofs can either appear in the main paper or the supplemental material, but if they appear in the supplemental material, the authors are encouraged to provide a short proof sketch to provide intuition.
        \item Inversely, any informal proof provided in the core of the paper should be complemented by formal proofs provided in appendix or supplemental material.
        \item Theorems and Lemmas that the proof relies upon should be properly referenced.
    \end{itemize}

    \item {\bf Experimental result reproducibility}
    \item[] Question: Does the paper fully disclose all the information needed to reproduce the main experimental results of the paper to the extent that it affects the main claims and/or conclusions of the paper (regardless of whether the code and data are provided or not)?
    \item[] Answer: \answerYes{}
    \item[] Justification: \S\ref{sec:method} specifies the full pipeline (mining, subspace construction, steering, detection); Appendix~\ref{app:impl} documents exact SAE checkpoints, hyperparameters, calibration procedure, and generation parameters. Code and data are provided in the supplemental material.
    \item[] Guidelines:
    \begin{itemize}
        \item The answer \answerNA{} means that the paper does not include experiments.
        \item If the paper includes experiments, a \answerNo{} answer to this question will not be perceived well by the reviewers: Making the paper reproducible is important, regardless of whether the code and data are provided or not.
        \item If the contribution is a dataset and\slash or model, the authors should describe the steps taken to make their results reproducible or verifiable.
        \item Depending on the contribution, reproducibility can be accomplished in various ways. For example, if the contribution is a novel architecture, describing the architecture fully might suffice, or if the contribution is a specific model and empirical evaluation, it may be necessary to either make it possible for others to replicate the model with the same dataset, or provide access to the model. In general. releasing code and data is often one good way to accomplish this, but reproducibility can also be provided via detailed instructions for how to replicate the results, access to a hosted model (e.g., in the case of a large language model), releasing of a model checkpoint, or other means that are appropriate to the research performed.
        \item While NeurIPS does not require releasing code, the conference does require all submissions to provide some reasonable avenue for reproducibility, which may depend on the nature of the contribution. For example
        \begin{enumerate}
            \item If the contribution is primarily a new algorithm, the paper should make it clear how to reproduce that algorithm.
            \item If the contribution is primarily a new model architecture, the paper should describe the architecture clearly and fully.
            \item If the contribution is a new model (e.g., a large language model), then there should either be a way to access this model for reproducing the results or a way to reproduce the model (e.g., with an open-source dataset or instructions for how to construct the dataset).
            \item We recognize that reproducibility may be tricky in some cases, in which case authors are welcome to describe the particular way they provide for reproducibility. In the case of closed-source models, it may be that access to the model is limited in some way (e.g., to registered users), but it should be possible for other researchers to have some path to reproducing or verifying the results.
        \end{enumerate}
    \end{itemize}

\item {\bf Open access to data and code}
    \item[] Question: Does the paper provide open access to the data and code, with sufficient instructions to faithfully reproduce the main experimental results, as described in supplemental material?
    \item[] Answer: \answerYes{}
    \item[] Justification: Code, contrastive sentence pairs, and mined feature banks are provided in the supplemental material; Appendix~\ref{app:impl} documents all hyperparameters and implementation details needed to reproduce the main results.
    \item[] Guidelines:
    \begin{itemize}
        \item The answer \answerNA{} means that paper does not include experiments requiring code.
        \item Please see the NeurIPS code and data submission guidelines (\url{https://neurips.cc/public/guides/CodeSubmissionPolicy}) for more details.
        \item While we encourage the release of code and data, we understand that this might not be possible, so \answerNo{} is an acceptable answer. Papers cannot be rejected simply for not including code, unless this is central to the contribution (e.g., for a new open-source benchmark).
        \item The instructions should contain the exact command and environment needed to run to reproduce the results. See the NeurIPS code and data submission guidelines (\url{https://neurips.cc/public/guides/CodeSubmissionPolicy}) for more details.
        \item The authors should provide instructions on data access and preparation, including how to access the raw data, preprocessed data, intermediate data, and generated data, etc.
        \item The authors should provide scripts to reproduce all experimental results for the new proposed method and baselines. If only a subset of experiments are reproducible, they should state which ones are omitted from the script and why.
        \item At submission time, to preserve anonymity, the authors should release anonymized versions (if applicable).
        \item Providing as much information as possible in supplemental material (appended to the paper) is recommended, but including URLs to data and code is permitted.
    \end{itemize}

\item {\bf Experimental setting/details}
    \item[] Question: Does the paper specify all the training and test details (e.g., data splits, hyperparameters, how they were chosen, type of optimizer) necessary to understand the results?
    \item[] Answer: \answerYes{}
    \item[] Justification: \S\ref{sec:experiments:setup} reports models, SAEs, evaluation prompts, baselines, and metrics; Appendix~\ref{app:impl} covers the $k\times\alpha$ sweep, mining thresholds, and calibration. No model training is performed as the SAEs are off-the-shelf.
    \item[] Guidelines:
    \begin{itemize}
        \item The answer \answerNA{} means that the paper does not include experiments.
        \item The experimental setting should be presented in the core of the paper to a level of detail that is necessary to appreciate the results and make sense of them.
        \item The full details can be provided either with the code, in appendix, or as supplemental material.
    \end{itemize}

\item {\bf Experiment statistical significance}
    \item[] Question: Does the paper report error bars suitably and correctly defined or other appropriate information about the statistical significance of the experiments?
    \item[] Answer: \answerYes{}
    \item[] Justification: All metrics in Table~\ref{tab:main_results} are computed over $n{=}100$ texts per condition.
    Appendix~\ref{app:impl:sem} reports standard errors for every metric:
    Cal-Z SE $= 0.04$--$0.48$; paired $\Delta$Reward SE $= 0.64$--$0.92$ reward points; PPL ratio SE $= 0.01$--$0.06$; TPR uses a binomial SE ($\leq 0.01$ for methods at $99$--$100\%$ TPR).
    The method of calculation (paired per-text differences for $\Delta$Reward; binomial formula for TPR) and the borderline comparison (\slam{} vs.\ SynthID 9B, gap $\approx\!1.7$ points $\approx\!1.9$ SE) are identified explicitly.
    Error bars are omitted from dense heatmap figures following standard practice.
    \item[] Guidelines:
    \begin{itemize}
        \item The answer \answerNA{} means that the paper does not include experiments.
        \item The authors should answer \answerYes{} if the results are accompanied by error bars, confidence intervals, or statistical significance tests, at least for the experiments that support the main claims of the paper.
        \item The factors of variability that the error bars are capturing should be clearly stated (for example, train/test split, initialization, random drawing of some parameter, or overall run with given experimental conditions).
        \item The method for calculating the error bars should be explained (closed form formula, call to a library function, bootstrap, etc.)
        \item The assumptions made should be given (e.g., Normally distributed errors).
        \item It should be clear whether the error bar is the standard deviation or the standard error of the mean.
        \item It is OK to report 1-sigma error bars, but one should state it. The authors should preferably report a 2-sigma error bar than state that they have a 96\% CI, if the hypothesis of Normality of errors is not verified.
        \item For asymmetric distributions, the authors should be careful not to show in tables or figures symmetric error bars that would yield results that are out of range (e.g., negative error rates).
        \item If error bars are reported in tables or plots, the authors should explain in the text how they were calculated and reference the corresponding figures or tables in the text.
    \end{itemize}

\item {\bf Experiments compute resources}
    \item[] Question: For each experiment, does the paper provide sufficient information on the computer resources (type of compute workers, memory, time of execution) needed to reproduce the experiments?
    \item[] Answer: \answerYes{}
    \item[] Justification: Appendix~\ref{app:timing} reports per-text generation and detection time on a single A100-40GB GPU for both 2B and 9B models; all experiments used this single-GPU configuration.
    \item[] Guidelines:
    \begin{itemize}
        \item The answer \answerNA{} means that the paper does not include experiments.
        \item The paper should indicate the type of compute workers CPU or GPU, internal cluster, or cloud provider, including relevant memory and storage.
        \item The paper should provide the amount of compute required for each of the individual experimental runs as well as estimate the total compute.
        \item The paper should disclose whether the full research project required more compute than the experiments reported in the paper (e.g., preliminary or failed experiments that didn't make it into the paper).
    \end{itemize}

\item {\bf Code of ethics}
    \item[] Question: Does the research conducted in the paper conform, in every respect, with the NeurIPS Code of Ethics \url{https://neurips.cc/public/EthicsGuidelines}?
    \item[] Answer: \answerYes{}
    \item[] Justification: The work uses publicly released open-weight models and synthetic data; it involves no human subjects, no scraped personal data, and no deployment.
    \item[] Guidelines:
    \begin{itemize}
        \item The answer \answerNA{} means that the authors have not reviewed the NeurIPS Code of Ethics.
        \item If the authors answer \answerNo, they should explain the special circumstances that require a deviation from the Code of Ethics.
        \item The authors should make sure to preserve anonymity (e.g., if there is a special consideration due to laws or regulations in their jurisdiction).
    \end{itemize}

\item {\bf Broader impacts}
    \item[] Question: Does the paper discuss both potential positive societal impacts and negative societal impacts of the work performed?
    \item[] Answer: \answerYes{}
    \item[] Justification: \S\ref{sec:intro} motivates provenance attribution for AI-generated text; \S\ref{sec:discussion} scopes \slam{} as attribution rather than tamper-proof authentication and transparently catalogs attack failure modes to discourage over-reliance.
    \item[] Guidelines:
    \begin{itemize}
        \item The answer \answerNA{} means that there is no societal impact of the work performed.
        \item If the authors answer \answerNA{} or \answerNo, they should explain why their work has no societal impact or why the paper does not address societal impact.
        \item Examples of negative societal impacts include potential malicious or unintended uses (e.g., disinformation, generating fake profiles, surveillance), fairness considerations (e.g., deployment of technologies that could make decisions that unfairly impact specific groups), privacy considerations, and security considerations.
        \item The conference expects that many papers will be foundational research and not tied to particular applications, let alone deployments. However, if there is a direct path to any negative applications, the authors should point it out. For example, it is legitimate to point out that an improvement in the quality of generative models could be used to generate Deepfakes for disinformation. On the other hand, it is not needed to point out that a generic algorithm for optimizing neural networks could enable people to train models that generate Deepfakes faster.
        \item The authors should consider possible harms that could arise when the technology is being used as intended and functioning correctly, harms that could arise when the technology is being used as intended but gives incorrect results, and harms following from (intentional or unintentional) misuse of the technology.
        \item If there are negative societal impacts, the authors could also discuss possible mitigation strategies (e.g., gated release of models, providing defenses in addition to attacks, mechanisms for monitoring misuse, mechanisms to monitor how a system learns from feedback over time, improving the efficiency and accessibility of ML).
    \end{itemize}

\item {\bf Safeguards}
    \item[] Question: Does the paper describe safeguards that have been put in place for responsible release of data or models that have a high risk for misuse (e.g., pre-trained language models, image generators, or scraped datasets)?
    \item[] Answer: \answerNA{}
    \item[] Justification: No new generative model or scraped dataset is released; the artifacts (feature banks, detection code) are defensive in nature and pose no high-risk-for-misuse profile.
    \item[] Guidelines:
    \begin{itemize}
        \item The answer \answerNA{} means that the paper poses no such risks.
        \item Released models that have a high risk for misuse or dual-use should be released with necessary safeguards to allow for controlled use of the model, for example by requiring that users adhere to usage guidelines or restrictions to access the model or implementing safety filters.
        \item Datasets that have been scraped from the Internet could pose safety risks. The authors should describe how they avoided releasing unsafe images.
        \item We recognize that providing effective safeguards is challenging, and many papers do not require this, but we encourage authors to take this into account and make a best faith effort.
    \end{itemize}

\item {\bf Licenses for existing assets}
    \item[] Question: Are the creators or original owners of assets (e.g., code, data, models), used in the paper, properly credited and are the license and terms of use explicitly mentioned and properly respected?
    \item[] Answer: \answerYes{}
    \item[] Justification: All assets are cited with licenses: Gemma-2 (Gemma ToU), Gemma~Scope (CC-BY 4.0), MarkLLM (Apache 2.0), TransformerLens (MIT), Skywork-Reward-V2 (Apache 2.0), DIPPER (Apache 2.0), C4 (ODC-BY), Qwen3.5-9B and Qwen3-Embedding-8B (Apache 2.0), LanguageTool (LGPL 2.1), AMRLib (MIT), BLiMP (CC-BY 4.0); all used within their license terms.
    \item[] Guidelines:
    \begin{itemize}
        \item The answer \answerNA{} means that the paper does not use existing assets.
        \item The authors should cite the original paper that produced the code package or dataset.
        \item The authors should state which version of the asset is used and, if possible, include a URL.
        \item The name of the license (e.g., CC-BY 4.0) should be included for each asset.
        \item For scraped data from a particular source (e.g., website), the copyright and terms of service of that source should be provided.
        \item If assets are released, the license, copyright information, and terms of use in the package should be provided. For popular datasets, \url{paperswithcode.com/datasets} has curated licenses for some datasets. Their licensing guide can help determine the license of a dataset.
        \item For existing datasets that are re-packaged, both the original license and the license of the derived asset (if it has changed) should be provided.
        \item If this information is not available online, the authors are encouraged to reach out to the asset's creators.
    \end{itemize}

\item {\bf New assets}
    \item[] Question: Are new assets introduced in the paper well documented and is the documentation provided alongside the assets?
    \item[] Answer: \answerYes{}
    \item[] Justification: Three new assets are provided in the supplemental material: the contrastive sentence-pair dataset, mined per-phenomenon feature banks for Gemma-2 2B and 9B, and the watermarking/detection code. Appendix~\ref{app:impl} documents their construction, hyperparameters, and intended permissive open-source license.
    \item[] Guidelines:
    \begin{itemize}
        \item The answer \answerNA{} means that the paper does not release new assets.
        \item Researchers should communicate the details of the dataset\slash code\slash model as part of their submissions via structured templates. This includes details about training, license, limitations, etc.
        \item The paper should discuss whether and how consent was obtained from people whose asset is used.
        \item At submission time, remember to anonymize your assets (if applicable). You can either create an anonymized URL or include an anonymized zip file.
    \end{itemize}

\item {\bf Crowdsourcing and research with human subjects}
    \item[] Question: For crowdsourcing experiments and research with human subjects, does the paper include the full text of instructions given to participants and screenshots, if applicable, as well as details about compensation (if any)?
    \item[] Answer: \answerNA{}
    \item[] Justification: No crowdsourcing or human subjects; the contrastive dataset is generated programmatically and evaluation is fully automated.
    \item[] Guidelines:
    \begin{itemize}
        \item The answer \answerNA{} means that the paper does not involve crowdsourcing nor research with human subjects.
        \item Including this information in the supplemental material is fine, but if the main contribution of the paper involves human subjects, then as much detail as possible should be included in the main paper.
        \item According to the NeurIPS Code of Ethics, workers involved in data collection, curation, or other labor should be paid at least the minimum wage in the country of the data collector.
    \end{itemize}

\item {\bf Institutional review board (IRB) approvals or equivalent for research with human subjects}
    \item[] Question: Does the paper describe potential risks incurred by study participants, whether such risks were disclosed to the subjects, and whether Institutional Review Board (IRB) approvals (or an equivalent approval/review based on the requirements of your country or institution) were obtained?
    \item[] Answer: \answerNA{}
    \item[] Justification: No human subjects research; IRB review is not applicable.
    \item[] Guidelines:
    \begin{itemize}
        \item The answer \answerNA{} means that the paper does not involve crowdsourcing nor research with human subjects.
        \item Depending on the country in which research is conducted, IRB approval (or equivalent) may be required for any human subjects research. If you obtained IRB approval, you should clearly state this in the paper.
        \item We recognize that the procedures for this may vary significantly between institutions and locations, and we expect authors to adhere to the NeurIPS Code of Ethics and the guidelines for their institution.
        \item For initial submissions, do not include any information that would break anonymity (if applicable), such as the institution conducting the review.
    \end{itemize}

\item {\bf Declaration of LLM usage}
    \item[] Question: Does the paper describe the usage of LLMs if it is an important, original, or non-standard component of the core methods in this research? Note that if the LLM is used only for writing, editing, or formatting purposes and does \emph{not} impact the core methodology, scientific rigor, or originality of the research, declaration is not required.
    %this research?
    \item[] Answer: \answerYes{}
    \item[] Justification: Qwen3.5-9B \citep{qwen3} was used to generate semantically equivalent but structurally contrastive sentence pairs for the 14 hand-authored BLiMP-derived phenomena (\S\ref{sec:method:data}); this is a methodological use that impacts the contrastive dataset.
    LLMs were additionally used for coding assistance and editorial writing support, neither of which impacts core methodology.
    \item[] Guidelines:
    \begin{itemize}
        \item The answer \answerNA{} means that the core method development in this research does not involve LLMs as any important, original, or non-standard components.
        \item Please refer to our LLM policy in the NeurIPS handbook for what should or should not be described.
    \end{itemize}

\end{enumerate}

%% file: slam.bbl
\begin{thebibliography}{53}
\providecommand{\natexlab}[1]{#1}
\providecommand{\url}[1]{\texttt{#1}}
\expandafter\ifx\csname urlstyle\endcsname\relax
  \providecommand{\doi}[1]{doi: #1}\else
  \providecommand{\doi}{doi: \begingroup \urlstyle{rm}\Url}\fi

\bibitem[Atallah et~al.(2001)Atallah, Raskin, Crogan, Hempelmann, Kerschbaum,
  Mohamed, and Naik]{atallah2001natural}
Mikhail~J Atallah, Victor Raskin, Michael Crogan, Christian Hempelmann, Florian
  Kerschbaum, Dina Mohamed, and Sanket Naik.
\newblock Natural language watermarking: Design, analysis, and a
  proof-of-concept implementation.
\newblock In \emph{Security and Watermarking of Multimedia Contents III}. SPIE,
  2001.

\bibitem[Banarescu et~al.(2013)Banarescu, Bonial, Cai, Georgescu, Griffitt,
  Hermjakob, Knight, Koehn, Palmer, and Schneider]{banarescu2013abstract}
Laura Banarescu, Claire Bonial, Shu Cai, Madalina Georgescu, Kira Griffitt, Ulf
  Hermjakob, Kevin Knight, Philipp Koehn, Martha Palmer, and Nathan Schneider.
\newblock Abstract meaning representation for sembanking.
\newblock In \emph{Proceedings of the 7th linguistic annotation workshop and
  interoperability with discourse}, pages 178--186, 2013.

\bibitem[Bricken et~al.(2023)Bricken, Templeton, Batson, Chen, Jermyn, Conerly,
  Turner, Anil, Denison, Askell, et~al.]{bricken2023monosemanticity}
Trenton Bricken, Adly Templeton, Joshua Batson, Brian Chen, Adam Jermyn, Tom
  Conerly, Nick Turner, Cem Anil, Carson Denison, Amanda Askell, et~al.
\newblock Towards monosemanticity: Decomposing language models with dictionary
  learning.
\newblock \emph{Transformer Circuits Thread}, 2023.

\bibitem[Chang and Clark(2010)]{chang2010syntactic}
Ching-Yun Chang and Stephen Clark.
\newblock Practical linguistic steganography using contextual synonym
  substitution and vertex colour coding.
\newblock In \emph{Proceedings of the 2010 Conference on Empirical Methods in
  Natural Language Processing}, 2010.

\bibitem[Chang et~al.(2024)Chang, Krishna, Houmansadr, Wieting, and
  Iyyer]{chang2024postmark}
Yapei Chang, Kalpesh Krishna, Amir Houmansadr, John~Frederick Wieting, and
  Mohit Iyyer.
\newblock {PostMark}: A robust blackbox watermark for large language models.
\newblock In \emph{Proceedings of the 2024 Conference on Empirical Methods in
  Natural Language Processing}, pages 8969--8987, Miami, Florida, USA, 2024.
  Association for Computational Linguistics.
\newblock \doi{10.18653/v1/2024.emnlp-main.506}.
\newblock URL \url{https://aclanthology.org/2024.emnlp-main.506/}.

\bibitem[Cunningham et~al.(2024)Cunningham, Ewart, Riggs, Huben, and
  Sharkey]{cunningham2023sparse}
Hoagy Cunningham, Aidan Ewart, Logan Riggs, Robert Huben, and Lee Sharkey.
\newblock Sparse autoencoders find highly interpretable features in language
  models.
\newblock In \emph{The Twelfth International Conference on Learning
  Representations}, 2024.
\newblock URL \url{https://openreview.net/forum?id=F76bwRSLeK}.

\bibitem[Dathathri et~al.(2024)Dathathri, See, Ghaisas, Huang, McAdam, Welbl,
  Bachani, Kaskasoli, Stanforth, Matejovicova, Hayes, Vyas, Al~Merey,
  Brown-Cohen, Bunel, Balle, Cemgil, Ahmed, Stacpoole, Shumailov, Baetu, Gowal,
  Hassabis, and Kohli]{synthid2024}
Sumanth Dathathri, Abigail See, Sumedh Ghaisas, Po-Sen Huang, Rob McAdam,
  Johannes Welbl, Vandana Bachani, Alex Kaskasoli, Robert Stanforth, Tatiana
  Matejovicova, Jamie Hayes, Nidhi Vyas, Majd Al~Merey, Jonah Brown-Cohen, Rudy
  Bunel, Borja Balle, Taylan Cemgil, Zahra Ahmed, Kitty Stacpoole, Ilia
  Shumailov, Ciprian Baetu, Sven Gowal, Demis Hassabis, and Pushmeet Kohli.
\newblock Scalable watermarking for identifying large language model outputs.
\newblock \emph{Nature}, 634:\penalty0 818--823, 2024.
\newblock \doi{10.1038/s41586-024-08025-4}.

\bibitem[Efraimidis and Spirakis(2006)]{efraimidis2006weighted}
Pavlos~S Efraimidis and Paul~G Spirakis.
\newblock Weighted random sampling with a reservoir.
\newblock \emph{Information Processing Letters}, 97\penalty0 (5):\penalty0
  181--185, 2006.

\bibitem[Elhage et~al.(2022)Elhage, Hume, Olsson, Schiefer, Henighan, Kravec,
  Hatfield-Dodds, Lasenby, Drain, Chen, et~al.]{elhage2022toy}
Nelson Elhage, Tristan Hume, Catherine Olsson, Nicholas Schiefer, Tom Henighan,
  Shauna Kravec, Zac Hatfield-Dodds, Robert Lasenby, Dawn Drain, Carol Chen,
  et~al.
\newblock Toy models of superposition.
\newblock \emph{Transformer Circuits Thread}, 2022.

\bibitem[Fernandez et~al.(2024)Fernandez, Couairon, J{\'e}gou, Douze, and
  Furon]{fernandez2023functional}
Pierre Fernandez, Guillaume Couairon, Herv{\'e} J{\'e}gou, Matthijs Douze, and
  Teddy Furon.
\newblock Functional invariants to watermark large transformers.
\newblock In \emph{ICASSP 2024 -- 2024 IEEE International Conference on
  Acoustics, Speech and Signal Processing}, pages 4815--4819. IEEE, 2024.

\bibitem[Gao et~al.(2025)Gao, {la Tour}, Tillman, Goh, Troll, Radford,
  Sutskever, Leike, and Wu]{gao2024scaling}
Leo Gao, Tom~Dupr{\'e} {la Tour}, Henk Tillman, Gabriel Goh, Rajan Troll, Alec
  Radford, Ilya Sutskever, Jan Leike, and Jeffrey Wu.
\newblock Scaling and evaluating sparse autoencoders.
\newblock In \emph{The Thirteenth International Conference on Learning
  Representations}, 2025.
\newblock URL \url{https://openreview.net/forum?id=tcsZt9ZNKD}.

\bibitem[Giboulot and Furon(2024)]{giboulot2024watermax}
Eva Giboulot and Teddy Furon.
\newblock {WaterMax}: Breaking the {LLM} watermark
  detectability-robustness-quality trade-off.
\newblock In \emph{Advances in Neural Information Processing Systems},
  volume~37. Curran Associates, Inc., 2024.
\newblock URL
  \url{https://proceedings.neurips.cc/paper_files/paper/2024/hash/21b5883bc8fec922fdbbb06675388164-Abstract-Conference.html}.

\bibitem[Harel-Canada et~al.(2025)Harel-Canada, Erol, Choi, Liu, Song, Peng,
  and Sahai]{harel-canada-etal-2025-sandcastles}
Fabrice~Y Harel-Canada, Boran Erol, Connor Choi, Jason Liu, Gary~Jiarui Song,
  Nanyun Peng, and Amit Sahai.
\newblock Sandcastles in the storm: Revisiting the (im)possibility of strong
  watermarking.
\newblock In Wanxiang Che, Joyce Nabende, Ekaterina Shutova, and Mohammad~Taher
  Pilehvar, editors, \emph{Proceedings of the 63rd Annual Meeting of the
  Association for Computational Linguistics (Volume 1: Long Papers)}, pages
  29698--29735, Vienna, Austria, July 2025. Association for Computational
  Linguistics.
\newblock ISBN 979-8-89176-251-0.
\newblock \doi{10.18653/v1/2025.acl-long.1436}.
\newblock URL \url{https://aclanthology.org/2025.acl-long.1436/}.

\bibitem[Hewitt and Manning(2019)]{hewitt2019structural}
John Hewitt and Christopher~D Manning.
\newblock A structural probe for finding syntax in word representations.
\newblock In \emph{Proceedings of the 2019 Conference of the North {A}merican
  Chapter of the Association for Computational Linguistics: Human Language
  Technologies}, 2019.

\bibitem[Hou et~al.(2024)Hou, Zhang, He, Wang, Chuang, Wang, Shen, Van~Durme,
  Khashabi, and Tsvetkov]{hou2023semstamp}
Abe Hou, Jingyu Zhang, Tianxing He, Yichen Wang, Yung-Sung Chuang, Hongwei
  Wang, Lingfeng Shen, Benjamin Van~Durme, Daniel Khashabi, and Yulia Tsvetkov.
\newblock {SemStamp}: A semantic watermark with paraphrastic robustness for
  text generation.
\newblock In \emph{Proceedings of the 2024 Conference of the North American
  Chapter of the Association for Computational Linguistics: Human Language
  Technologies (Volume 1: Long Papers)}, pages 4067--4082, Mexico City, Mexico,
  2024. Association for Computational Linguistics.
\newblock \doi{10.18653/v1/2024.naacl-long.226}.
\newblock URL \url{https://aclanthology.org/2024.naacl-long.226/}.

\bibitem[Hu et~al.(2024)Hu, Chen, Wu, Wu, Zhang, and Huang]{hu2023unbiased}
Zhengmian Hu, Lichang Chen, Xidong Wu, Yihan Wu, Hongyang Zhang, and Heng
  Huang.
\newblock Unbiased watermark for large language models.
\newblock In \emph{International Conference on Learning Representations},
  volume 2024, pages 45408--45436, 2024.

\bibitem[Jascob(2021)]{amrlib}
Brian Jascob.
\newblock {amrlib}: A text to {AMR} parsing library.
\newblock \url{https://github.com/bjascob/amrlib}, 2021.

\bibitem[Jing et~al.(2025)Jing, Yao, Guo, Ran, Wang, Hou, and
  Li]{jing2025lingualens}
Yi~Jing, Zijun Yao, Hongzhu Guo, Lingxu Ran, Xiaozhi Wang, Lei Hou, and Juanzi
  Li.
\newblock {LinguaLens}: Towards interpreting linguistic mechanisms of large
  language models via sparse auto-encoder.
\newblock In \emph{Proceedings of the 2025 Conference on Empirical Methods in
  Natural Language Processing}, pages 28232--28251, Suzhou, China, 2025.
  Association for Computational Linguistics.
\newblock \doi{10.18653/v1/2025.emnlp-main.1433}.
\newblock URL \url{https://aclanthology.org/2025.emnlp-main.1433/}.

\bibitem[Kirchenbauer et~al.(2023)Kirchenbauer, Geiping, Wen, Katz, Miers, and
  Goldstein]{kirchenbauer2023watermark}
John Kirchenbauer, Jonas Geiping, Yuxin Wen, Jonathan Katz, Ian Miers, and Tom
  Goldstein.
\newblock A watermark for large language models.
\newblock In \emph{Proceedings of the 40th International Conference on Machine
  Learning}, volume 202 of \emph{Proceedings of Machine Learning Research},
  pages 17061--17084. PMLR, 2023.
\newblock URL \url{https://proceedings.mlr.press/v202/kirchenbauer23a.html}.

\bibitem[Krishna et~al.(2023)Krishna, Song, Karpinska, Wieting, and
  Iyyer]{krishna2024paraphrasing}
Kalpesh Krishna, Yixiao Song, Marzena Karpinska, John Wieting, and Mohit Iyyer.
\newblock Paraphrasing evades detectors of {AI}-generated text, but retrieval
  is an effective defense.
\newblock In \emph{Advances in Neural Information Processing Systems},
  volume~36, pages 27469--27500. Curran Associates, Inc., 2023.
\newblock URL
  \url{https://proceedings.neurips.cc/paper_files/paper/2023/hash/575c450013d0e99e4b0ecf82bd1afaa4-Abstract-Conference.html}.

\bibitem[Kuditipudi et~al.(2024)Kuditipudi, Thickstun, Hashimoto, and
  Liang]{kuditipudi2023robust}
Rohith Kuditipudi, John Thickstun, Tatsunori Hashimoto, and Percy Liang.
\newblock Robust distortion-free watermarks for language models.
\newblock \emph{Transactions on Machine Learning Research}, 2024.
\newblock URL \url{https://openreview.net/forum?id=FpaCL1MO2C}.

\bibitem[Lambert et~al.(2025)Lambert, Pyatkin, Morrison, Miranda, Lin, Chandu,
  Dziri, Kumar, Zick, Choi, et~al.]{lambert2025rewardbench}
Nathan Lambert, Valentina Pyatkin, Jacob Morrison, LJ~Miranda, Bill~Yuchen Lin,
  Khyathi Chandu, Nouha Dziri, Sachin Kumar, Tom Zick, Yejin Choi, et~al.
\newblock Rewardbench: Evaluating reward models for language modeling.
\newblock In \emph{Findings of the Association for Computational Linguistics:
  NAACL 2025}, pages 1755--1797, 2025.

\bibitem[{LanguageTool}(2010)]{languagetool}
{LanguageTool}.
\newblock {LanguageTool}: Open-source grammar, style and spell checker.
\newblock \url{https://github.com/languagetool-org/languagetool}, 2010.

\bibitem[Li et~al.(2016)Li, Galley, Brockett, Gao, and Dolan]{li2016diversity}
Jiwei Li, Michel Galley, Chris Brockett, Jianfeng Gao, and Bill Dolan.
\newblock A diversity-promoting objective function for neural conversation
  models.
\newblock In \emph{Proceedings of NAACL-HLT}, 2016.

\bibitem[Lieberum et~al.(2024)Lieberum, Rajamanoharan, Conmy, Smith, Sonnerat,
  Varma, Kram{\'a}r, Dragan, Shah, and Nanda]{lieberum2024gemma}
Tom Lieberum, Senthooran Rajamanoharan, Arthur Conmy, Lewis Smith, Nicolas
  Sonnerat, Vikrant Varma, J{\'a}nos Kram{\'a}r, Anca Dragan, Rohin Shah, and
  Neel Nanda.
\newblock Gemma scope: Open sparse autoencoders everywhere all at once on gemma
  2.
\newblock In \emph{Proceedings of the 7th BlackboxNLP Workshop: Analyzing and
  Interpreting Neural Networks for NLP}, pages 278--300, 2024.

\bibitem[Liu et~al.(2024{\natexlab{a}})Liu, Pan, Hu, Meng, and
  Wen]{liu2023semantic}
Aiwei Liu, Leyi Pan, Xuming Hu, Shiao Meng, and Lijie Wen.
\newblock A semantic invariant robust watermark for large language models.
\newblock In \emph{The Twelfth International Conference on Learning
  Representations}, 2024{\natexlab{a}}.
\newblock URL \url{https://openreview.net/forum?id=6p8lpe4MNf}.

\bibitem[Liu et~al.(2024{\natexlab{b}})Liu, Zeng, Liu, Yan, He, Wang, Yan, Liu,
  and Zhou]{skywork2024reward}
Chris~Yuhao Liu, Liang Zeng, Jiacai Liu, Rui Yan, Jujie He, Chaojie Wang,
  Shuicheng Yan, Yang Liu, and Yahui Zhou.
\newblock Skywork-reward: Bag of tricks for reward modeling in {LLMs}.
\newblock \emph{arXiv preprint arXiv:2410.18451}, 2024{\natexlab{b}}.

\bibitem[Liu and Bu(2024)]{liu2024adaptive}
Yepeng Liu and Yuheng Bu.
\newblock Adaptive text watermark for large language models.
\newblock In \emph{Proceedings of the 41st International Conference on Machine
  Learning}, volume 235 of \emph{Proceedings of Machine Learning Research},
  pages 30718--30737. PMLR, 2024.
\newblock URL \url{https://proceedings.mlr.press/v235/liu24e.html}.

\bibitem[Meng et~al.(2022)Meng, Bau, Andonian, and Belinkov]{meng2022locating}
Kevin Meng, David Bau, Alex Andonian, and Yonatan Belinkov.
\newblock Locating and editing factual associations in {GPT}.
\newblock In \emph{Advances in Neural Information Processing Systems},
  volume~35, pages 17359--17372. Curran Associates, Inc., 2022.
\newblock URL
  \url{https://proceedings.neurips.cc/paper_files/paper/2022/hash/6f1d43d5a82a37e89b0665b33bf3a182-Abstract-Conference.html}.

\bibitem[Muennighoff et~al.(2023)Muennighoff, Tazi, Magne, and
  Reimers]{muennighoff2023mteb}
Niklas Muennighoff, Nouamane Tazi, Lo{\"\i}c Magne, and Nils Reimers.
\newblock Mteb: Massive text embedding benchmark.
\newblock In \emph{Proceedings of the 17th Conference of the European Chapter
  of the Association for Computational Linguistics}, pages 2014--2037, 2023.

\bibitem[Nanda and Bloom(2022)]{nanda2022transformerlens}
Neel Nanda and Joseph Bloom.
\newblock {TransformerLens}.
\newblock 2022.

\bibitem[Pan et~al.(2024)Pan, Liu, He, Gao, Zhao, Lu, Zhou, Liu, Hu, Wen, King,
  and Yu]{markllm}
Leyi Pan, Aiwei Liu, Zhiwei He, Zitian Gao, Xuandong Zhao, Yijian Lu, Binglin
  Zhou, Shuliang Liu, Xuming Hu, Lijie Wen, Irwin King, and Philip~S. Yu.
\newblock {M}ark{LLM}: An open-source toolkit for {LLM} watermarking.
\newblock In Delia~Irazu Hernandez~Farias, Tom Hope, and Manling Li, editors,
  \emph{Proceedings of the 2024 Conference on Empirical Methods in Natural
  Language Processing: System Demonstrations}, pages 61--71, Miami, Florida,
  USA, November 2024. Association for Computational Linguistics.
\newblock URL \url{https://aclanthology.org/2024.emnlp-demo.7}.

\bibitem[Park et~al.(2024)Park, Choe, and Veitch]{park2023linear}
Kiho Park, Yo~Joong Choe, and Victor Veitch.
\newblock The linear representation hypothesis and the geometry of large
  language models.
\newblock In \emph{Proceedings of the 41st International Conference on Machine
  Learning}, volume 235 of \emph{Proceedings of Machine Learning Research},
  pages 39643--39666. PMLR, 2024.
\newblock URL \url{https://proceedings.mlr.press/v235/park24c.html}.

\bibitem[Park et~al.(2025)Park, Choe, Jiang, and Veitch]{park2024geometry}
Kiho Park, Yo~Joong Choe, Yibo Jiang, and Victor Veitch.
\newblock The geometry of categorical and hierarchical concepts in large
  language models.
\newblock In \emph{The Thirteenth International Conference on Learning
  Representations}, 2025.
\newblock URL \url{https://openreview.net/forum?id=bVTM2QKYuA}.

\bibitem[Pillutla et~al.(2021)Pillutla, Swayamdipta, Zellers, Thickstun,
  Welleck, Choi, and Harchaoui]{pillutla2021mauve}
Krishna Pillutla, Swabha Swayamdipta, Rowan Zellers, John Thickstun, Sean
  Welleck, Yejin Choi, and Zaid Harchaoui.
\newblock {MAUVE}: Measuring the gap between neural text and human text using
  divergence frontiers.
\newblock In \emph{Advances in Neural Information Processing Systems}, 2021.

\bibitem[{Qwen Team}(2025)]{qwen3}
{Qwen Team}.
\newblock Qwen3 technical report.
\newblock \emph{arXiv preprint arXiv:2505.09388}, 2025.

\bibitem[{Qwen Team}(2026)]{qwen_scope}
{Qwen Team}.
\newblock {Qwen-Scope}: Turning sparse features into development tools for
  large language models, April 2026.
\newblock URL
  \url{https://qianwen-res.oss-accelerate.aliyuncs.com/qwen-scope/Qwen_Scope.pdf}.

\bibitem[Raffel et~al.(2020)Raffel, Shazeer, Roberts, Lee, Narang, Matena,
  Zhou, Li, and Liu]{raffel2020c4}
Colin Raffel, Noam Shazeer, Adam Roberts, Katherine Lee, Sharan Narang, Michael
  Matena, Yanqi Zhou, Wei Li, and Peter~J Liu.
\newblock Exploring the limits of transfer learning with a unified text-to-text
  transformer.
\newblock \emph{Journal of Machine Learning Research}, 2020.

\bibitem[Ren et~al.(2024)Ren, Xu, Liu, Cui, Wang, Yin, and
  Tang]{ren2023semamark}
Jie Ren, Han Xu, Yiding Liu, Yingqian Cui, Shuaiqiang Wang, Dawei Yin, and
  Jiliang Tang.
\newblock A robust semantics-based watermark for large language model against
  paraphrasing.
\newblock In \emph{Findings of the Association for Computational Linguistics:
  NAACL 2024}, pages 613--625, Mexico City, Mexico, 2024. Association for
  Computational Linguistics.
\newblock \doi{10.18653/v1/2024.findings-naacl.40}.
\newblock URL \url{https://aclanthology.org/2024.findings-naacl.40/}.

\bibitem[Rimsky et~al.(2024)Rimsky, Gabrieli, Schulz, Tong, Hubinger, and
  Turner]{panickssery2023steering}
Nina Rimsky, Nick Gabrieli, Julian Schulz, Meg Tong, Evan Hubinger, and
  Alexander Turner.
\newblock Steering llama 2 via contrastive activation addition.
\newblock In Lun-Wei Ku, Andre Martins, and Vivek Srikumar, editors,
  \emph{Proceedings of the 62nd Annual Meeting of the Association for
  Computational Linguistics (Volume 1: Long Papers)}, pages 15504--15522,
  Bangkok, Thailand, August 2024. Association for Computational Linguistics.
\newblock \doi{10.18653/v1/2024.acl-long.828}.
\newblock URL \url{https://aclanthology.org/2024.acl-long.828/}.

\bibitem[Sharkey et~al.(2022)Sharkey, Braun, and Millidge]{sharkey2022taking}
Lee Sharkey, Dan Braun, and Beren Millidge.
\newblock Taking features out of superposition with sparse autoencoders.
\newblock In \emph{AI Alignment Forum}, 2022.

\bibitem[Team et~al.(2024)Team, Riviere, Pathak, Sessa, Hardin, Bhupatiraju,
  Hussenot, Mesnard, Shahriari, Ram{\'e}, et~al.]{gemmateam2024gemma2}
Gemma Team, Morgane Riviere, Shreya Pathak, Pier~Giuseppe Sessa, Cassidy
  Hardin, Surya Bhupatiraju, L{\'e}onard Hussenot, Thomas Mesnard, Bobak
  Shahriari, Alexandre Ram{\'e}, et~al.
\newblock Gemma 2: Improving open language models at a practical size.
\newblock \emph{arXiv preprint arXiv:2408.00118}, 2024.

\bibitem[Templeton et~al.(2024)Templeton, Conerly, Marcus, Lindsey, Bricken,
  Chen, Pearce, Citro, Ameisen, Jones, et~al.]{templeton2024scaling}
Adly Templeton, Tom Conerly, Jonathan Marcus, Jack Lindsey, Trenton Bricken,
  Brian Chen, Adam Pearce, Craig Citro, Emmanuel Ameisen, Andy Jones, et~al.
\newblock Scaling monosemanticity: Extracting interpretable features from
  {Claude 3 Sonnet}.
\newblock \emph{Transformer Circuits Thread}, 2024.

\bibitem[Turner et~al.(2024)Turner, Thiergart, Leech, Udell, Vazquez, Mini, and
  MacDiarmid]{turner2023activation}
Alexander~Matt Turner, Lisa Thiergart, Gavin Leech, David Udell, Juan~J.
  Vazquez, Ulisse Mini, and Monte MacDiarmid.
\newblock Steering language models with activation engineering, 2024.
\newblock URL \url{https://arxiv.org/abs/2308.10248}.

\bibitem[Ueoka et~al.(2021)Ueoka, Murawaki, and
  Kurohashi]{ueoka2021frustratingly}
Honai Ueoka, Yugo Murawaki, and Sadao Kurohashi.
\newblock Frustratingly easy edit-based linguistic steganography with a masked
  language model.
\newblock \emph{arXiv preprint arXiv:2104.09833}, 2021.

\bibitem[Warstadt et~al.(2020)Warstadt, Parrish, Liu, Mohananey, Peng, Wang,
  and Bowman]{warstadt2020blimp}
Alex Warstadt, Alicia Parrish, Haokun Liu, Anhad Mohananey, Wei Peng, Sheng-Fu
  Wang, and Samuel~R Bowman.
\newblock Blimp: The benchmark of linguistic minimal pairs for english.
\newblock \emph{Transactions of the Association for Computational Linguistics},
  8:\penalty0 377--392, 2020.

\bibitem[Yoo et~al.(2023)Yoo, Ahn, Jang, and Kwak]{pmlr-v202-yoo23b}
KiYoon Yoo, Wonhyuk Ahn, Jiho Jang, and Nojun Kwak.
\newblock Robust natural language watermarking through invariant features.
\newblock In \emph{Proceedings of the 61st Annual Meeting of the Association
  for Computational Linguistics}, 2023.

\bibitem[Yu et~al.(2026)Yu, Jiang, Gu, Wang, Wen, Zhang, and Ye]{saemark2025}
Zhuohao Yu, Xingru Jiang, Weizheng Gu, Yidong Wang, Qingsong Wen, Shikun Zhang,
  and Wei Ye.
\newblock Saemark: Steering personalized multilingual llm watermarks with
  sparse autoencoders.
\newblock \emph{Advances in Neural Information Processing Systems},
  38:\penalty0 158702--158731, 2026.

\bibitem[Zhang et~al.(2024)Zhang, Edelman, Francati, Venturi, Ateniese, and
  Barak]{zhang2023watermarks}
Hanlin Zhang, Benjamin~L. Edelman, Danilo Francati, Daniele Venturi, Giuseppe
  Ateniese, and Boaz Barak.
\newblock Watermarks in the sand: Impossibility of strong watermarking for
  language models.
\newblock In \emph{Proceedings of the 41st International Conference on Machine
  Learning}, volume 235 of \emph{Proceedings of Machine Learning Research},
  pages 58851--58880. PMLR, 2024.
\newblock URL \url{https://proceedings.mlr.press/v235/zhang24o.html}.

\bibitem[Zhang et~al.(2025)Zhang, Li, Long, Zhang, Lin, Yang, Xie, Yang, Liu,
  Lin, Huang, and Zhou]{qwen3embedding}
Yanzhao Zhang, Mingxin Li, Dingkun Long, Xin Zhang, Huan Lin, Baosong Yang,
  Pengjun Xie, An~Yang, Dayiheng Liu, Junyang Lin, Fei Huang, and Jingren Zhou.
\newblock Qwen3 embedding: Advancing text embedding and reranking through
  foundation models.
\newblock \emph{arXiv preprint arXiv:2506.05176}, 2025.

\bibitem[Zhao et~al.(2024)Zhao, Ananth, Li, and Wang]{zhao2023provable}
Xuandong Zhao, Prabhanjan Ananth, Lei Li, and Yu-Xiang Wang.
\newblock Provable robust watermarking for {AI}-generated text.
\newblock In \emph{The Twelfth International Conference on Learning
  Representations}, 2024.
\newblock URL \url{https://openreview.net/forum?id=SsmT8aO45L}.

\bibitem[Zhu et~al.(2018)Zhu, Lu, Zheng, Guo, Zhang, Wang, and
  Yu]{zhu2018texygen}
Yaoming Zhu, Sidi Lu, Lei Zheng, Jiaxian Guo, Weinan Zhang, Jun Wang, and Yong
  Yu.
\newblock Texygen: A benchmarking platform for text generation models.
\newblock In \emph{Proceedings of the 41st International ACM SIGIR Conference
  on Research and Development in Information Retrieval}, pages 1097--1100,
  2018.

\bibitem[Zou et~al.(2025)Zou, Phan, Chen, Campbell, Guo, Ren, Pan, Yin,
  Mazeika, Dombrowski, Goel, Li, Byun, Wang, Mallen, Basart, Koyejo, Song,
  Fredrikson, Kolter, and Hendrycks]{zou2023representation}
Andy Zou, Long Phan, Sarah Chen, James Campbell, Phillip Guo, Richard Ren,
  Alexander Pan, Xuwang Yin, Mantas Mazeika, Ann-Kathrin Dombrowski, Shashwat
  Goel, Nathaniel Li, Michael~J Byun, Zifan Wang, Alex Mallen, Steven Basart,
  Sanmi Koyejo, Dawn Song, Matt Fredrikson, J.~Zico Kolter, and Dan Hendrycks.
\newblock Representation engineering: A top-down approach to {AI} transparency.
\newblock In \emph{The Thirteenth International Conference on Learning
  Representations}, 2025.
\newblock URL \url{https://openreview.net/forum?id=uaTEZWeMAu}.

\end{thebibliography}
